\documentclass[10pt,journal,compsoc]{IEEEtran}

\usepackage{graphicx}
\usepackage[flushleft]{threeparttable}
\usepackage{colortbl}
\usepackage{amsmath}
\usepackage{amsthm}
\usepackage{amssymb}
\usepackage{enumitem}
\usepackage{bm}
\usepackage{bbm}
\usepackage{multirow}
\usepackage{color}
\usepackage{tabu,comment}
\usepackage{array, eucal}
\usepackage[ruled,vlined]{algorithm2e}
\ifCLASSOPTIONcompsoc
  \usepackage[nocompress]{cite}
\else
  \usepackage{cite}
\fi

\ifCLASSOPTIONcompsoc
  \usepackage[caption=false,font=footnotesize,labelfont=sf,textfont=sf]{subfig}
\else
  \usepackage[caption=false,font=footnotesize]{subfig}
\fi

\usepackage[pagebackref=true,breaklinks=true,letterpaper=true,colorlinks,citecolor=blue,linkcolor=blue,bookmarks=false]{hyperref}

\newtheorem{remark}{Remark}
\graphicspath{{./Figures/}}
\newcommand{\mat}[1]{\mbox{\boldsymbol{$#1$}}}
\definecolor{mygray}{gray}{.9}

\newcommand{\squishlist}{
	\begin{list}{$\bullet$}
		{ \setlength{\itemsep}{0pt}
			\setlength{\parsep}{2pt}
			\setlength{\topsep}{2pt}
			\setlength{\partopsep}{0pt}
			\setlength{\leftmargin}{1em}
			\setlength{\labelwidth}{1em}
			\setlength{\labelsep}{0.5em} } }
	\newcommand{\squishend}{
\end{list}  }

\begin{document}
\title{Index Networks}

\author{Hao~Lu,
        Yutong~Dai,
        Chunhua~Shen, and Songcen~Xu%
\IEEEcompsocitemizethanks{\IEEEcompsocthanksitem H. Lu, Y. Dai and C. Shen are with The University of Adelaide, SA 5005, Australia. Corresponding author: C. Shen.\protect\\
E-mail: {firstname.lastname}@adelaide.edu.au
\IEEEcompsocthanksitem S. Xu is with Noah's Ark Lab, Huawei Technologies.\protect\\
}%
}

\markboth{Manuscript. V1: August 2019; V2: April 2020}%
{Lu \MakeLowercase{\textit{et al.}}: Index Networks}

\IEEEtitleabstractindextext{%
\begin{abstract}

We show that existing upsampling operators can be unified using the notion of the index function. This notion is inspired by an observation in the decoding process of deep image matting where indices-guided unpooling can often recover boundary details considerably better than other upsampling operators such as bilinear interpolation. By viewing the indices as a function of the feature map, we introduce the concept of `learning to index', and present a novel index-guided encoder-decoder framework where indices are learned adaptively from data and are used to guide downsampling and upsampling stages, without extra training supervision. At the core of this framework is a new learnable module, termed Index Network (IndexNet), which dynamically generates indices conditioned on the feature map. IndexNet can be used as a plug-in applicable
to almost all convolutional networks that have coupled downsampling and upsampling stages, enabling the networks to dynamically capture variations of local patterns. In particular, we instantiate, investigate five families of IndexNet, highlight their superiority in delivering spatial information over other upsampling operators with experiments on synthetic data, and demonstrate their effectiveness on four dense prediction tasks, including image matting, image denoising, semantic segmentation, and monocular depth estimation. Code and models are available at: \url{https://git.io/IndexNet}.

\end{abstract}

\begin{IEEEkeywords}
Upsampling Operators, Dynamic Networks,
Image Denoising, Semantic Segmentation,
Image Matting, Depth Estimation
\end{IEEEkeywords}}

\maketitle

\IEEEdisplaynontitleabstractindextext

\IEEEpeerreviewmaketitle

\IEEEraisesectionheading{\section{Introduction}\label{sec:introduction}}

\IEEEPARstart{U}{psampling} is an essential stage for dense prediction tasks using deep convolutional neural networks (CNNs). The frequently used upsampling operators include transposed convolution~\cite{zeiler2014visualizing,long2015fully}, unpooling~\cite{badrinarayanan2017segnet}, \textit{periodic shuffling}~\cite{shi2016real} (a.k.a.\  depth-to-space), and naive interpolation~\cite{lin2017refine,chen18v3} followed by convolution. These operators, however, are not general-purpose designs and often exhibit different behaviors in different tasks.

\begin{figure}[!t]
	\captionsetup{font=small,singlelinecheck=true}
	\begin{center}
		\includegraphics[width=\linewidth]{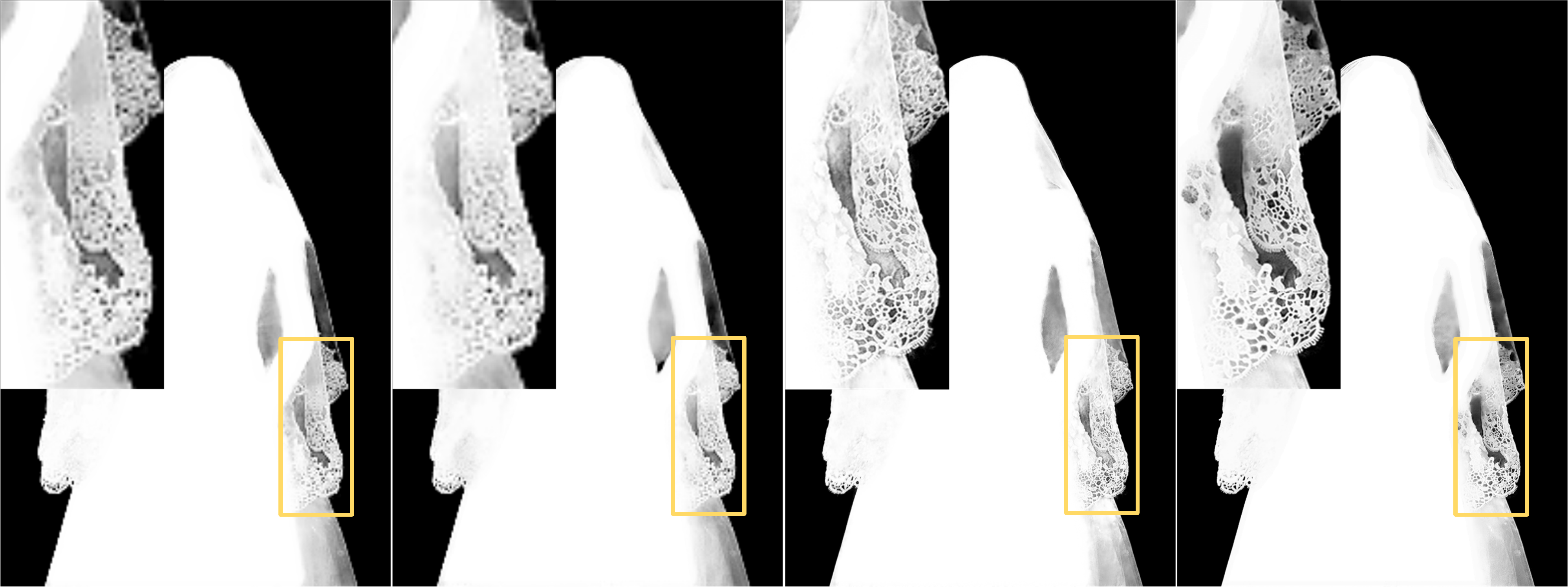}\vspace{-10pt}
	\end{center}
	\caption{Alpha mattes of different models for the task of image matting. From left to right, Deeplabv3+ \cite{chen18v3}, RefineNet \cite{lin2017refine}, Deep Matting
	\cite{xu2017deep} and IndexNet (Ours). Bilinear upsampling tends to fail to recover subtle details, while unpooling and our learned upsampling operator can produce much clear mattes with good local contrast.
	}
	\label{fig:alpha_mattes}
\end{figure}

The widely-adopted upsampling operator in semantic segmentation and depth estimation is bilinear interpolation, while unpooling is less popular. A reason might be that the feature map generated by max unpooling is sparse, while the bilinearly interpolated feature map has dense and consistent representations for local regions (compared to the feature map before interpolation). This is particularly true for semantic segmentation and depth estimation where pixels in a region often share the same class label or have similar depth. However, we observe that bilinear interpolation can perform significantly worse than unpooling in boundary-sensitive tasks such as image matting. A fact is that the leading deep image matting model~\cite{xu2017deep} largely borrows the design from the SegNet method \cite{badrinarayanan2017segnet}, where unpooling was first introduced. When adapting other state-of-the-art segmentation models, such as DeepLabv3+~\cite{chen18v3} and RefineNet~\cite{lin2017refine}, to this task, we observe that they tend to fail to recover boundary details (Fig.~\ref{fig:alpha_mattes}). A plausible explanation is that, compared to the bilinearly upsampled feature map, unpooling uses max-pooling indices to guide upsampling. Since boundaries in the shallow layers usually have the maximum responses, indices extracted from these responses record the boundary locations. The feature map projected by the indices thus shows improved boundary delineation.

We thus believe that different upsampling operators may
exhibit
different characteristics, and we expect a specific behavior of the upsampling operator when dealing with specific image content for a particular vision task. A question of interest is: \textit{Can we design a generic operator to upsample feature maps that better predict boundaries and regions simultaneously?} A key observation of this work is that unpooling, bilinear interpolation or other \textit{upsampling operators are some forms of index functions}. For example, the nearest neighbor interpolation of a point is equivalent to allocating indices of one to its neighbor and then map the value of the point. In this sense, indices are models~\cite{kraska2018case}, therefore indices can be modeled and learned.

In this work, \textit{we model indices as a function of the local feature map and learn index functions to implement upsampling within deep CNNs.} In particular, we present a novel index-guided encoder-decoder framework, which naturally generalizes models like SegNet. Instead of using max-pooling and unpooling, we introduce indexed pooling and indexed upsampling where downsampling and upsampling are guided by learned indices. The indices are generated \textit{dynamically } conditioned on the feature map and are learned using a fully convolutional network, termed IndexNet, without extra
supervision needed. IndexNet is a highly flexible module. It can be applied to almost all convolutional networks that have coupled downsampling and upsampling stages. Compared to the fixed $\max$ function or bilinear interpolation, learned index functions show potentials for simultaneous boundary and region delineation.

IndexNet is a high-level concept and represents a broad family of networks modeling the so-called index function. In this work, we instantiate and investigate five families of IndexNet. Different designs correspond to different assumptions.
We compare the behavior of IndexNet with existing upsampling operators and demonstrate its superiority in delivering spatial information. We show that IndexNet can be incorporated into many CNNs to benefit a number of visual tasks, for instance:
i) \textit{image matting}: our MobileNetv2-based~\cite{sandler2018mobilenetv2} model with IndexNet exhibits at least $16.1\%$ improvement against the VGG-16-based DeepMatting baseline~\cite{xu2017deep} on the Composition-1k matting dataset; by visualizing learned indices, the indices automatically learn to capture the boundaries and textural patterns;
ii) \textit{image denoising}: a modified DnCNN model with IndexNet can achieve performance comparable to the baseline DnCNN~\cite{zhang2017beyond} that has no downsampling stage on the BSD68 and Set12 datasets~\cite{roth2009fields}, thus reducing the computational cost and memory consumption significantly;
iii) \textit{semantic segmentation}: consistently improved performance is observed when SegNet~\cite{badrinarayanan2017segnet} is equipped with IndexNet on the SUN RGB-D dataset~\cite{song2015sun}; and
iv) \textit{monocular depth estimation}: IndexNet also improves the performance of a recent light-weight FastDepth model on the NYUDv2 dataset~\cite{silberman2012indoor}, with negligible extra computation cost.

We make the following main contributions:
\begin{itemize}
    \item We present a unified perspective of existing upsampling operators with the notion of the index function;
	\item We introduce Index Networks---a novel family of networks that can be included into standard CNNs to provide dynamic, adaptive downsampling and upsampling capabilities;
	to the best of our knowledge, IndexNet is one of the first attempts towards the design of generic upsampling operators;
	\item We instantiate, and investigate five designs of IndexNet and demonstrate their effectiveness on four vision tasks.

\end{itemize}

A preliminary conference version of this work appeared in~\cite{hao2019indexnet}. We extend~\cite{hao2019indexnet} by i) further investigating two light-weight IndexNets;
ii) comparing properties and computational complexity of IndexNets;
iii) presenting a taxonomy of upsampling operators with the notion of guided upsampling and blind upsampling;
and iv) designing image reconstruction experiments on synthetic data to compare two upsampling paradigms in recovering spatial information. Besides image matting in~\cite{hao2019indexnet}, we now v) apply IndexNets to a few more vision tasks including image denoising, semantic segmentation and monocular depth estimation and report extensive experimental results, not only between index networks but also across different vision tasks.

\section{Literature Review}

We review upsampling operators and a closely-related group of networks%
: dynamic networks.

{\bf Upsampling in Deep Networks.}
Compared with other components in the design of deep networks, downsampling and upsampling of feature maps are relatively less studied. Since learning a CNN without sacrificing the spatial resolution is computationally expensive and memory intensive, and suffers from limited receptive fields, downsampling operators are common choices, such as strided convolution and max/average pooling. To recover the resolution, upsampling is thus an essential stage for almost all dense prediction tasks. This poses a fundamental question: \textit{What is the principal approach to recover the resolution of a downsampled feature map (decoding)}. A few
upsampling operators are proposed. The \textit{deconvolution} operator, a.k.a.\ transposed convolution, was initially used in~\cite{zeiler2014visualizing} to visualize convolutional activations and introduced to semantic segmentation~\cite{long2015fully}, but this operator
sometimes can be
harmful
due to its behavior in producing checkerboard artifacts~\cite{odena2016deconvolution}. To avoid this, a suggestion is the ``resize+convolution'' paradigm, which has currently become the standard configuration in state-of-the-art semantic segmentation models~\cite{chen18v3,lin2017refine}. Apart from these, \textit{perforate}~\cite{osendorfer2014image} and \textit{unpooling}~\cite{badrinarayanan2017segnet} generate sparse indices to guide upsampling. The indices are able to capture and keep boundary information, but one issue is that the two operators can induce much sparsity after upsampling. Convolutional layers with large filter sizes must follow for densification. In addition, \textit{periodic shuffling} ($\mathcal{PS}$) was introduced in~\cite{shi2016real} as a fast and memory-efficient upsampling operator for image super-resolution. $\mathcal{PS}$ recovers resolution by rearranging the feature map of size $H\times W\times Cr^2$ to $rH\times rW\times C$. It is also used in some segmentation models~\cite{yang2019deeperlab}.

Our work is primarily inspired by the unpooling operator~\cite{badrinarayanan2017segnet}. We remark that, it is important to extract spatial information before its loss during downsampling, and more importantly, to use stored information during upsampling. Unpooling shows a simple and effective use case, while we believe that there is much room to improve. Here we show that unpooling is a special form of index function, and we can learn an index function beyond unpooling.

We notice that concurrent work of~\cite{jiaqi2019carafe} also pursues the idea of data-dependent upsampling and proposes an universal upsampling operator termed CARAFE.
Although the idea is similar, IndexNet is different from CARAFE in several aspects. First, CARAFE does not associate upsampling with the notion of the index function.
Second, the kernels used in CARAFE are generated conditioned on decoder features, while IndexNet builds upon encoder features, so the generated indices can also be used to guide downsampling. Third, CARAFE can be viewed as one of our investigated index networks---holistic index networks, but with different upsampling kernels and normalization strategies.
We further compare IndexNet with CARAFE in Section~\ref{ssec:comparison} and in experiments.

\textbf{Dynamic Networks.}
If considering the dynamic property of IndexNet, {IndexNet} shares similarity with an interesting group of networks---dynamic networks. Dynamic networks are often implemented with adaptive modules
to extend the modeling capabilities of CNNs. These networks share the following characteristics. The output
is
\textit{dynamic}, conditioned on the input feature map.
Since dynamic networks are learnable modules, they are
\textit{generic} in the sense that they can be used as building blocks in many network architectures. They are also %
\textit{flexible} to allow modifications according to target
tasks.

\textit{Spatial Transformer Networks (STNs)}~\cite{jaderberg2015spatial}. STN allows explicit manipulation of spatial transformation within the network. It achieves this by regressing transformation parameters $\theta$ with a side-branch network. A spatially-transformed output is then produced by a sampler parameterized by $\theta$. This results in a holistic transformation of the feature map.
The dynamic nature of STN is reflected by the fact that, given different inputs, the inferred $\theta$ is different, allowing to learn some forms of invariance to translation, scale, rotation, etc.

\textit{Dynamic Filter Networks (DFNs)}~\cite{jia2016dynamic}. DFN implements a
filter generating network to dynamically generate kernel filter parameters
Compared to conventional filter parameters that
stay fixed during inference, filter parameters in DFN are dynamic and sample-specific.

\textit{Deformable Convolutional Networks (DCNs)}~\cite{dai2017deformable}. DCNs introduce deformable transformation into convolution. The key idea is to predict offsets for convolutional kernels. With offsets, convolution can be executed on irregular sampling grids, %
enabling adaptive
manipulation of the receptive field.

\textit{Attention Networks}~\cite{mnih2014recurrent}. Attention networks are a broad family of networks that use attention mechanisms. The mechanisms introduce multiplicative interactions between the inferred attention map and the feature map. In computer vision, attention mechanisms are usually referred to spatial attention~\cite{wang2017residual}, channel attention~\cite{hu2018squeeze} or both~\cite{woo2018cbam}.
These network modules are widely applied in CNNs to force the network focusing on specific regions and therefore to refine feature maps.
Essentially,
attention is about feature selection.
No attentional module has been designed to deal with the downsampling/upsampling stage.

In contrast to above dynamic networks, IndexNet
focuses on
upsampling, rather than manipulating filters or refining features.
Akin to
dynamic networks above, the dynamics in IndexNet
also has a
physical definition---indices. Such a definition also closely relates to attention networks.
Later we show that the downsampling and upsampling operators used with IndexNet can, to some extent, be viewed as attentional operators.
Indeed, max-pooling indices are a form of hard attention. It is worth noting that, despite that IndexNet in its current implementation
may closely relate to attention, it
focuses on
upsampling rather than refining feature maps.
IndexNet also shares the other characteristics mentioned above. It is implemented in a convolutional side-branch network, is trained without extra supervision and is generic and flexible.
We demonstrate
its effectiveness on four dense prediction tasks and five variants of IndexNet.

\begin{figure*}[!tb]
	\captionsetup{font=small,singlelinecheck=true}
	\setlength{\abovecaptionskip}{10pt}
	\centering
	\includegraphics[width=5in,angle=0]{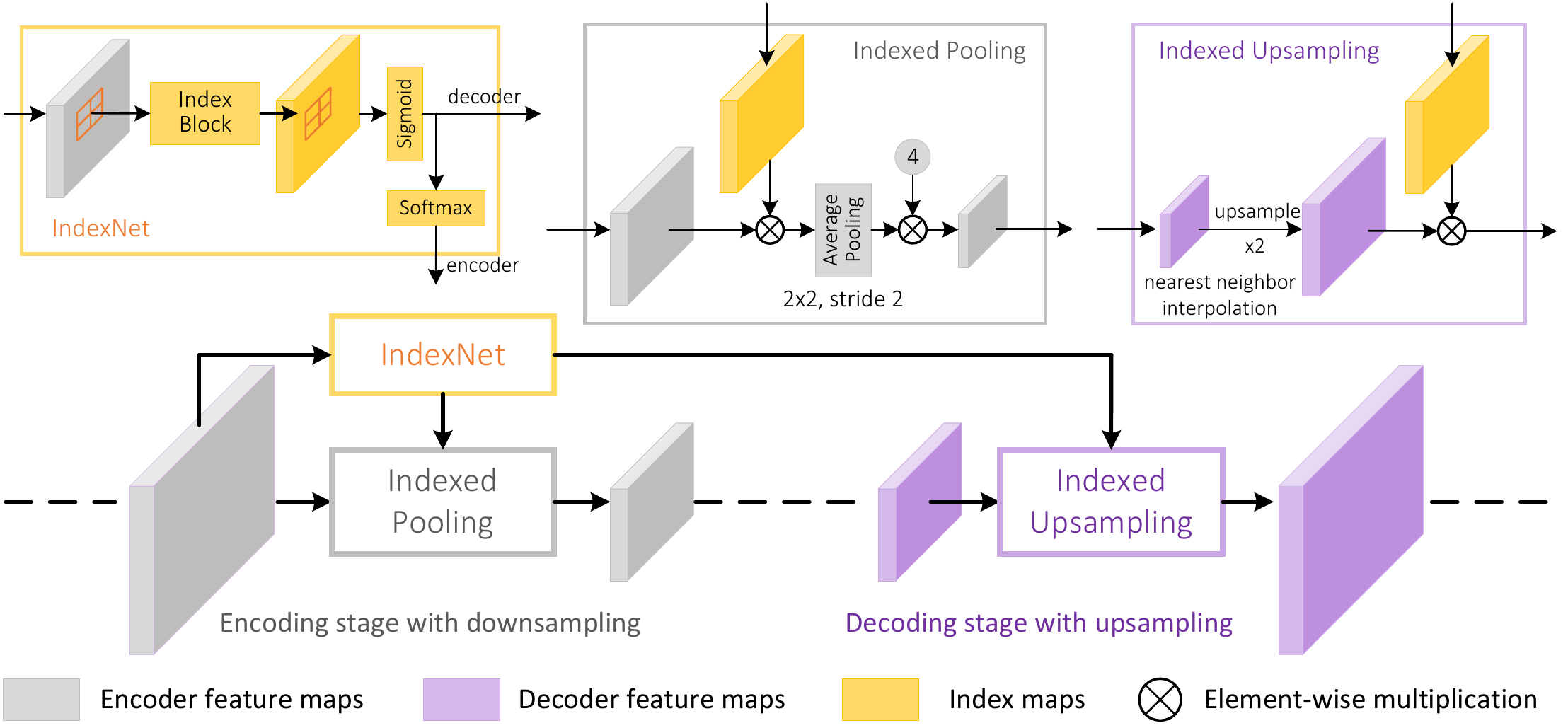}\vspace{-10pt}
	\caption{The index-guided encoder-decoder framework. The proposed {IndexNet} dynamically predicts indices for individual local regions, conditioned  on the input local feature map itself. The predicted indices are further used to guide the downsampling in the encoding stage and the upsampling in the corresponding decoding stage.}
	\label{fig:framework}
\end{figure*}

\section{An Indexing Perspective of Upsampling}

With the argument that upsampling operators are index functions, here we offer a unified indexing perspective of upsampling operators. The unpooling operator is straightforward. We can define its index function in a $k\times k$ local region as an indicator function
\begin{equation}
I_{max}(x)=\mathbbm{1}(x=\max (\mat X))\,,x\in\mat X \,,
\end{equation}
where $\mat X\in\mathbb{R}^{k\times k}$.
$ \mathbbm{1}( \cdot ) $ is the indicator function with output being a binary matrix.
Similarly, if one extracts indices from average pooling, its index function takes the form
\begin{equation}
I_{avg}(x)=\mathbbm{1}(x\in\mat X)\,.
\end{equation}
If further using $I_{avg}(x)$ during upsampling, it is equivalent to the nearest neighbor interpolation. As for the bilinear interpolation and deconvolution operators, their index functions have an identical form
\begin{equation}\label{eq:dconv}
I_{bilinear/dconv}(x)=\mat W\otimes\mathbbm{1}(x\in\mat X)\,,
\end{equation}
where $\mat W$ is the weight/filter of the same size as $\mat X$, and $\otimes$ denotes the element-wise multiplication. The difference is that, $\mat W$ is learned in deconvolution but %
predefined
in bilinear interpolation. Indeed bilinear interpolation has been shown to be a special case of deconvolution~\cite{long2015fully}. Note that, in this case, the index function generates soft indices. The sense of index for the $\mathcal{PS}$ operator~\cite{shi2016real} is
also
clear, because the rearrangement of the feature map %
is an indexing process. Considering $\mathcal{PS}$ a tensor $\mathcal{Z}$ of size $1\times1\times r^2$ to a matrix $\mat Z$ of size $r\times r$, the index function can be expressed by the one-hot encoding
\begin{equation}
I^l_{ps}(x)=\mathbbm{1}(x= \mathcal{Z}_l)\,,l=1,...,r^2\,,
\end{equation}
such that $\mat Z_{m,n}=\mathcal{Z}[I^l_{ps}(x)]$, where $m=1,...,r$, $n=1,...,r$, and $l=(r-1) \cdot m+n$. $\mathcal{Z}_l$ denotes the $l$-th element of $\mathcal{Z}$.
Similar notation applies to $\mat Z_{m,n}$.

Since upsampling operators can be unified by the notion of the index function, it is
plausible
to %
ask
whether one can learn an index function to dynamically capture local spatial patterns.

\section{Learning to Index, to Pool, and to Upsample}

Before introducing
the
designs of IndexNet, we first present
the general idea
about how learned indices may be used in downsampling and upsampling with a new index-guided encoder-decoder framework. Our framework is a
generalization of SegNet, as
illustrated in Fig.~\ref{fig:framework}. For ease of exposition,
let us
assume the downsampling and upsampling rates
to be
$2$, and the pooling operator
to use a
kernel size of $2\times2$. The IndexNet module dynamically generates indices given the feature map. The proposed indexed pooling and indexed upsampling operators further receive generated indices to guide downsampling and upsampling, respectively. In practice, multiple such modules can be combined and used analogous to the max pooling layers for every downsampling and upsampling stage.

\textit{IndexNet} models the index as a function of the feature map $\mathcal{X}\in\mathbb{R}^{H\times W\times C}$. Given $\mathcal{X}$, it generates two index maps for downsampling and upsampling, respectively. An important concept for the index is that an index can either be represented in a natural order, e.g., 1, 2, 3, ..., or be represented in a logical form, i.e., 0, 1, 0, ...,
meaning that an index map can be used as a mask. This is exactly how we use the index map in downsampling/upsampling. The predicted index shares the same %
definition of the index in computer science, except that we generate \textit{soft} indices for smooth optimization, i.e., for any index $i$, $i\in[0,1]$.

IndexNet consists of a predefined index block and two index normalization layers. An index block can simply be a heuristically defined function, e.g., a $\max$  function, or more generally, a parameterized function such as
neural network. In this work,
we use a fully convolutional network to be
the index block.
More details are presented
in Sections~\ref{subsec:holistic_networks} and
\ref{subsec:depthwise_networks}.
Note  that the index maps sent to the encoder and decoder are normalized differently. The decoder index map only goes through a \textit{sigmoid} function such that for any predicted index $i\in(0,1)$. As for the encoder index map, indices of each local region $L$ are further normalized by a \textit{softmax} function such that $\sum_{i\in{L}}i=1$.
The
second normalization
guarantees the magnitude consistency of the feature map after downsampling.

\textit{Indexed Pooling} ($\mathcal{IP}$)
performs
downsampling using generated indices. Given a local region $E\in\mathbb{R}^{k\times k}$, $\mathcal{IP}$ calculates a weighted sum of activations and corresponding indices over $E$ as $\mathcal{IP}(E)=\sum_{x\in E}I(x)x$, where $I(x)$ is the index of $x$. It is easy to
see
that max pooling and average pooling are special cases of $\mathcal{IP}$. In practice, this operator can be easily implemented with an element-wise multiplication between the feature map and the index map, an average pooling layer, and a multiplication of a constant used to compensate the effect of averaging, as instantiated in Fig.~\ref{fig:framework}. The current implementation is equivalent to $2\times2$ stride-2 convolution with dynamic kernels, but is more efficient than explicit on-the-fly
kernel generation.

\textit{Indexed Upsampling} ($\mathcal{IU}$) is the inverse operator of $\mathcal{IP}$. $\mathcal{IU}$ upsamples $d\in\mathbb{R}^{1\times 1}$ that spatially corresponds to $E$ taking the same indices into account. Let $I\in\mathbb{R}^{k\times k}$ be the local index map formed by $I(x)$s, $\mathcal{IU}$ upsamples $d$ as $\mathcal{IU}(d)=I\otimes D$, where $\otimes$ denotes the element-wise multiplication, and $D$ is of the same size as $I$ and is upsampled from $d$ with the nearest neighbor interpolation. $\mathcal{IU}$ also relates to deconvolution, but an important difference between $\mathcal{IU}$ and deconvolution is that, deconvolution applies a fixed kernel to all local regions (even if the kernel is learned), while $\mathcal{IU}$ upsamples different regions with different kernels (indices).

\begin{figure}[!tb]
	\captionsetup{font=small,singlelinecheck=true}
	\setlength{\abovecaptionskip}{10pt}
	\centering
	\includegraphics[width=3in,angle=0]{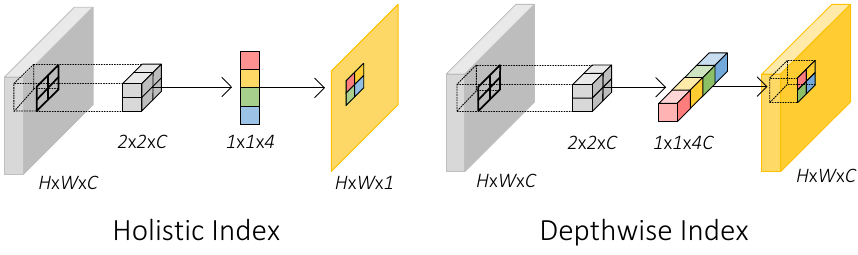}\vspace{-10pt}
	\caption{Conceptual differences between holistic and depthwise index.}
	\label{fig:holistic_depthwise}
\end{figure}

\subsection{Index Networks}

\begin{figure}[!tb]
	\captionsetup{font=small,singlelinecheck=true}
	\setlength{\abovecaptionskip}{10pt}
	\centering
	\includegraphics[width=2.43in,angle=0]{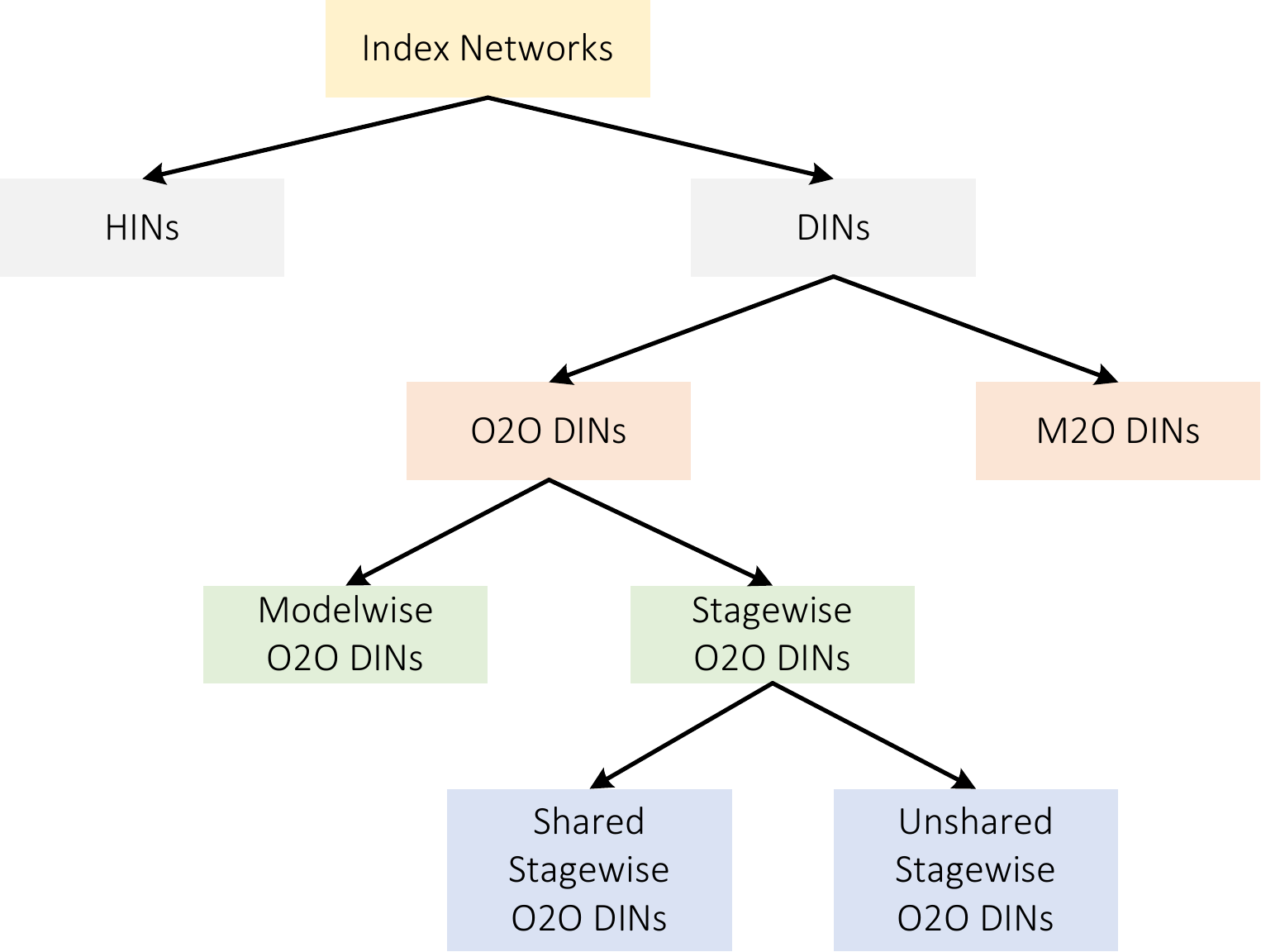}\vspace{-10pt}
	\caption{A taxonomy of proposed index networks.}
	\label{fig:indexnet_taxonomy}
\end{figure}

Here we present a taxonomy of proposed index networks. According to the shape of the output index map, index networks can be first categorized into two branches: \textit{holistic index networks} (HINs) and \textit{depthwise (separable) index networks} (DINs). Their conceptual differences are shown in Fig.~\ref{fig:holistic_depthwise}. HINs learn an index function $I(\mathcal{X}):\mathbb{R}^{H\times W\times C}\rightarrow\mathbb{R}^{H\times W\times1}$. In this case, all channels of the feature map share a holistic index map. By contrast, DINs learn an index function $I(\mathcal{X}):\mathbb{R}^{H\times W\times C}\rightarrow\mathbb{R}^{H\times W \times C}$, where the index map is of the same size as the feature map.

\begin{figure}[!tb]
	\captionsetup{font=small,singlelinecheck=true}
	\setlength{\abovecaptionskip}{10pt}
	\centering
	\includegraphics[width=3.4in,angle=0]{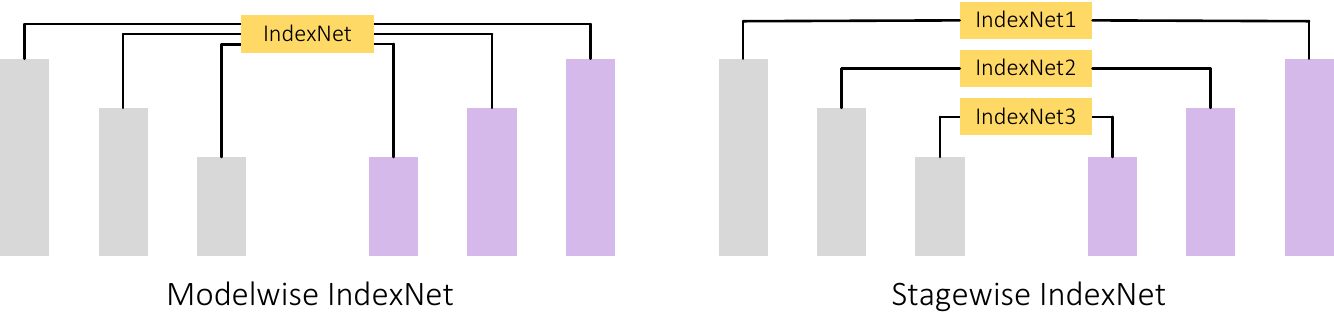}\vspace{-10pt}
	\caption{Modelwise IndexNet vs.\  stagewise IndexNet.}
	\label{fig:modelwise_stagewise}
\end{figure}

Since the index map generated by DINs can correspond to individual slices of the feature map, we can incorporate further assumptions into DINs to simplify the designs. If assuming that each slice of the index map only relates to its corresponding slice of the feature map, this %
is the
One-to-One (O2O) assumption and \textit{O2O DINs}. If each slice of the index map relates to all channels of the feature map, this leads to Many-to-One (M2O) assumption and \textit{M2O DINs}. In O2O DINs, one can further consider sharing IndexNet. In the most simplified case, the same IndexNet can be applied to every slice of the feature map and can be shared across different downsampling/upsampling stages, like the $\max$ function. We name this IndexNet \textit{Modelwise O2O DINs}. If IndexNet is stage-specific, i.e., only sharing indices in individual stages, we call this IndexNet \textit{Shared Stagewise O2O DINs}. Finally, without sharing any parameter (each feature slice has its specific index function), we
obtain
the standard design, %
termed
\textit{Unshared Stagewise O2O DINs}. Fig.~\ref{fig:indexnet_taxonomy} %
shows
the tree diagram of these index networks. The difference between modelwise IndexNet and stagewise IndexNet is also shown in Fig.~\ref{fig:modelwise_stagewise}. Notice that, HINs and M2O DINs are both stagewise.

With the taxonomy, %
we investigate five families of IndexNet. Each family can be designed to have either linear mappings or nonlinear mappings,
as we discuss next.

\subsection{Holistic Index Networks}\label{subsec:holistic_networks}

Recall that HINs learn an index function $I(\mathcal{X}):\mathbb{R}^{H\times W\times C}\rightarrow\mathbb{R}^{H\times W\times1}$. A naive design choice is to assume a linear mapping between the feature map and the index map.

\begin{figure}[!tb]
	\captionsetup{font=small,singlelinecheck=true}
	\setlength{\abovecaptionskip}{10pt}
	\centering
	\includegraphics[width=2.4in,angle=0]{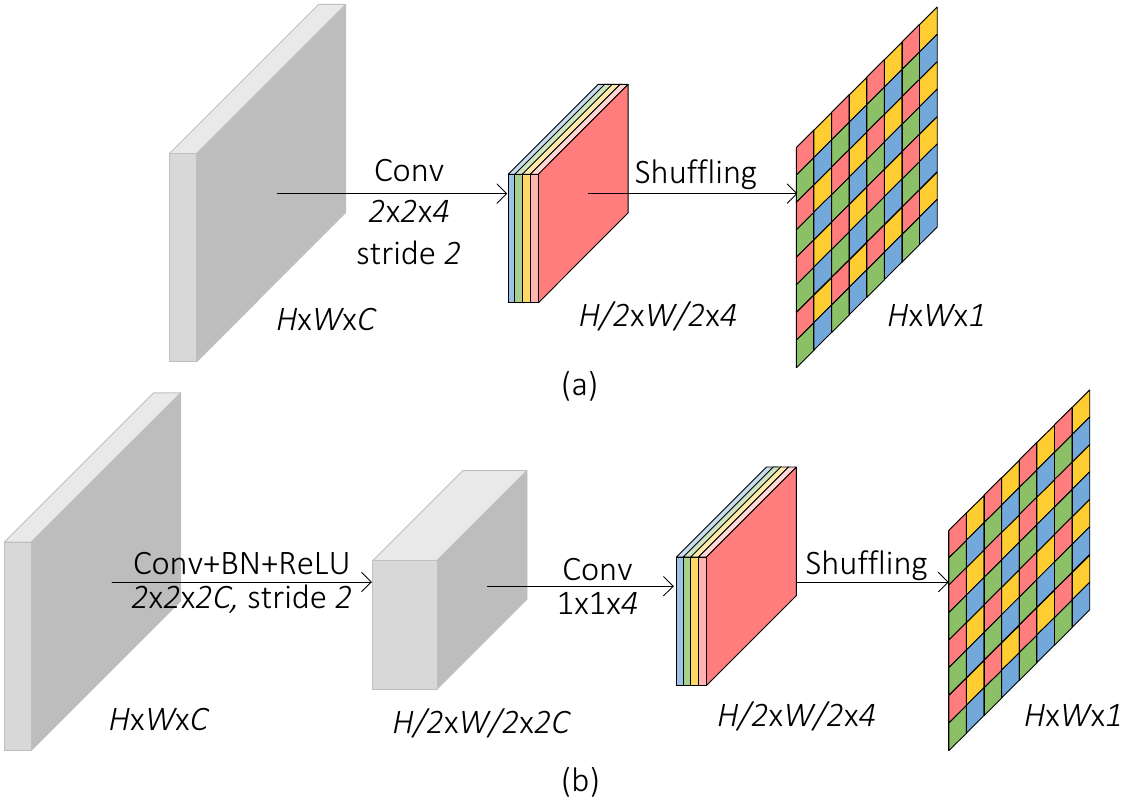}\vspace{-10pt}
	\caption{Holistic index networks. (a) a linear index network; (b) a nonlinear index network.}
	\label{fig:holistic_networks}
\end{figure}

\textit{Linear HINs.} An example is shown in Fig.~\ref{fig:holistic_networks}(a). The network is implemented in a fully convolutional network. It first applies stride-$2$
$2\times2$ convolution (assuming that the downsampling rate is $2$) to the feature map of size $H\times W\times C$, generating a concatenated index map of size $H/2\times W/2\times 4$. Each slice of the index map ($H/2\times W/2\times 1$) is designed to correspond to the indices of a certain position of all local regions, e.g., the top-left corner of all $2\times 2$ regions. The network finally applies a $\mathcal{PS}$-like shuffling operator to rearrange the index map to the size of $H\times W\times 1$.

In many situations, %
a linear relationship is not sufficient. For example,
a linear function even cannot approximate the max function.
Thus, the second design choice is to
introduce
nonlinearity into the network.

\textit{Nonlinear HINs.} Fig.~\ref{fig:holistic_networks}(b) illustrates a nonlinear HIN where the feature map is first projected to a map of size $H/2\times W/2\times 2C$, followed by a batch normalization layer and a ReLU function for nonlinear mappings. We then use point-wise convolution to reduce the channel dimension to an indices-compatible size. The
remaining
transformations follow its linear counterpart.

\subsection{Depthwise Index Networks}\label{subsec:depthwise_networks}

In DINs, we seek $I(\mathcal{X}):\mathbb{R}^{H\times W\times C}\rightarrow\mathbb{R}^{H\times W \times C}$, i.e., each spatial index corresponds to each spatial activation. As aforementioned, this type of networks further has two different high-level design strategies that correspond to two different assumptions.

\subsubsection{One-to-One Depthwise Index Networks}

O2O assumption assumes that each slice of the index map only relates to its corresponding slice of the feature map. It can be denoted by a local index function $l(\mathcal{X}): \mathbb{R}^{k\times k\times1}\rightarrow\mathbb{R}^{k\times k\times1}$, where $k$ denotes the size of the local region. Since the local index function operates on individual feature slices, we can design whether different feature slices share the same local index function. Such a weight sharing strategy
can
be applied at a modelwise level or at a stagewise level, which
leads to
the following designs of O2O DINs:
\begin{enumerate}
	\item \textit{Modelwise O2O DINs}: the model only has a unique index function that is shared by all feature slices, even in different downsampling and upsampling stages. This is the most light-weight design;
	\item \textit{Shared Stagewise O2O DINs}: the index function is also shared by feature slices, but every stage has stage-specific IndexNet. This design is also light-weight;
	\item \textit{Unshared Stagewise O2O DINs}: even in the same stage, different feature slices have distinct index functions.
\end{enumerate}

\begin{figure}[!tb]
	\captionsetup{font=small,singlelinecheck=true}
	\setlength{\abovecaptionskip}{10pt}
	\centering
	\includegraphics[width=3.2in,angle=0]{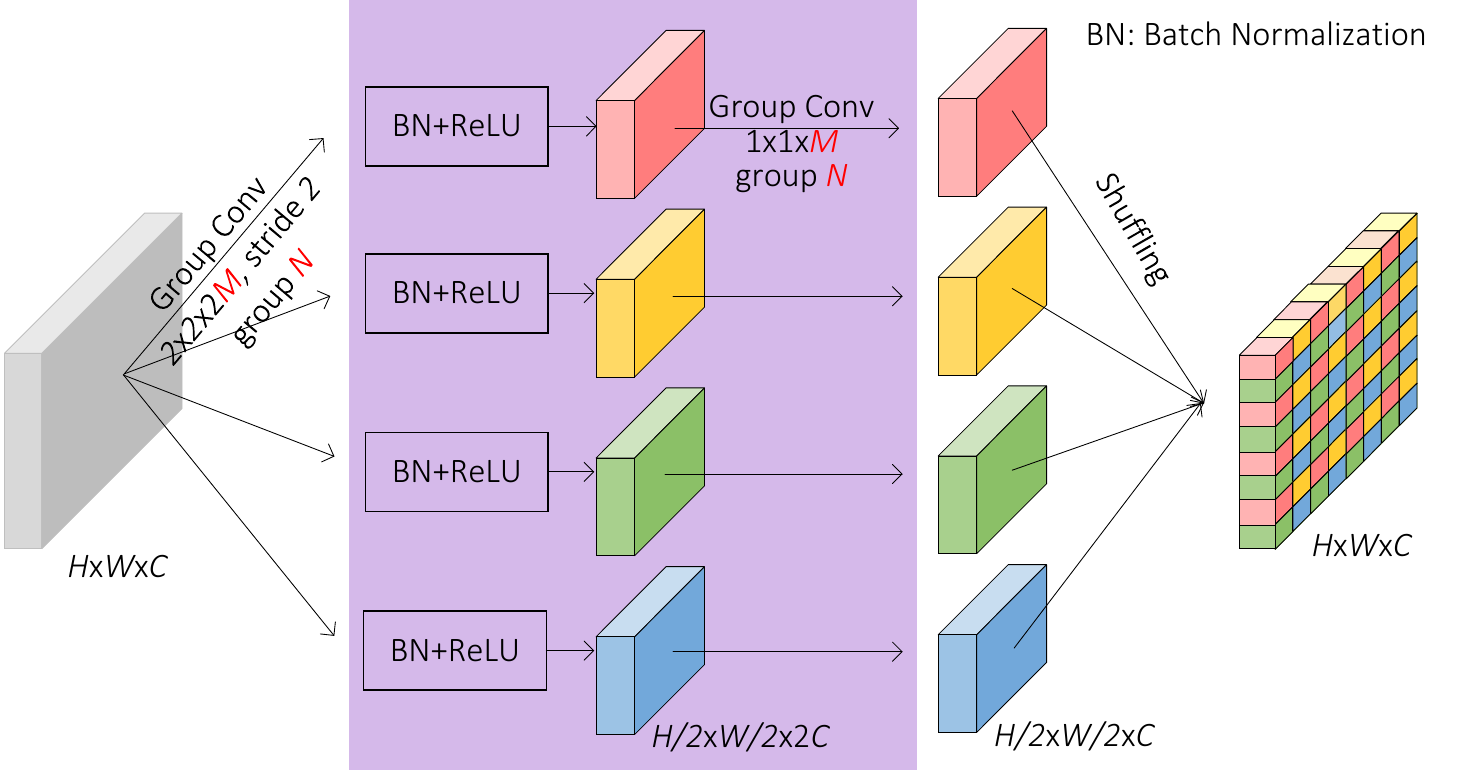}\vspace{-10pt}
	\caption{Depthwise index networks. $M=1,N=1$ for Modelwise O2O DINs and Shared Stagewise O2O DINs; $M=C,N=C$ for Unshared Stagewise O2O DINs; and $M=C,N=1$ for the M2O DINs. The masked modules are invisible to linear networks.}
	\label{fig:depthwise_networks}
\end{figure}

Similar to HINs, DINs can also be designed to have linear/nonlinear modeling ability. Fig.~\ref{fig:depthwise_networks} shows an example when $k=2$. Note that, in contrast to HINs, DINs follow a multi-column architecture. Each column is responsible for predicting indices specific to a certain spatial location of all local regions. We implement DINs with group convolutions.

\textit{Linear O2O DINs.}  According to Fig.~\ref{fig:depthwise_networks}, the feature map first goes through four parallel convolutional layers with the same kernel size. Modelwise O2O DINs and Shared Stagewise O2O DINs only
use
a kernel size of $2\times2\times 1$, a stride of $2$, and $1$ group, while Unshared Stagewise O2O DINs has a kernel size of $2\times2\times C$, a stride of $2$, and $C$ groups. One can simply reshape the feature map, i.e., reshaping $H\times W \times C$ to be $C \times H\times W \times 1$, to enable a $2\times2\times 1$ kernel operating on each $H\times W \times 1$ feature slice, respectively. All O2O DINs lead to four downsampled feature maps of size $H/2\times W/2 \times C$. The final index map of size $H\times W \times C$ is composed from the four feature maps by shuffling and rearrangement. Note that the parameters of four columns are not shared.

\textit{Nonlinear O2O DINs.} Nonlinear DINs can be easily modified from linear DINs by inserting four extra convolutional layers. Each of them is followed by a batch normalization (BN) layer and a ReLU unit, as shown in Fig.~\ref{fig:depthwise_networks}. The rest remains the same as the linear DINs.

\subsubsection{Many-to-One Depthwise Index Networks}
M2O assumption assumes that all feature slices have contributions to each index slice. The local index function is defined by $l(\mathcal{X}): \mathbb{R}^{k\times k\times C}\rightarrow\mathbb{R}^{k\times k\times1}$. Compared to O2O DINs, the only difference in implementation is the use of standard convolution instead of group convolution, i.e., $M=C,N=1$ in Fig.~\ref{fig:depthwise_networks}.

\subsection{Property and Model Complexity}

Both HINs and DINs have merits and drawbacks. Here we discuss
some important properties of IndexNet.
We also present an analysis of computational complexity.

\begin{remark}
	Index maps generated by HINs and used by the $\mathcal{IP}$ and $\mathcal{IU}$ operators
	are
	related to spatial attention.
\end{remark}

The holistic index map is shared by all feature slices, which means
that
the index map is required to be expanded to the size of $H\times W\times C$ when feeding into $\mathcal{IP}$ and $\mathcal{IU}$.
This index map
can
be thought as a collection of local attention maps~\cite{mnih2014recurrent} applied to individual local spatial regions. In this case, the $\mathcal{IP}$ and $\mathcal{IU}$ operators can also be referred to as ``attentional pooling'' and ``attentional upsampling''. However, it should %
be noted that spatial attention has no pooling or upsampling
operators
like $\mathcal{IP}$ and $\mathcal{IU}$.

\begin{remark}
	HINs are more flexible than DINs and
	more friendly for decoder design.
\end{remark}

Since the holistic index map is expandable, the decoder feature map does not need to forcibly increase/reduce its dimensionality to fit the shape of the index map during upsampling. This gives much flexibility for decoder design, while it is not the case for DINs.

\begin{remark}
	The number of parameters in Modelwise O2O DINs and Shared Stagewise O2O DINs is independent of the dimensionality of feature maps.
\end{remark}

No matter how large the model capacity is or how wide the feature channels are, the number of parameters in Modelwise O2O DINs remains at a constant level, and that in Shared Stagewise O2O DINs is only proportional to the number of downsampling/upsampling stages. This is
desirable as the number of parameters introduced by IndexNet is not significant.
However, these two types of IndexNet may be limited to capture sophisticated local patterns.

\begin{remark}
	M2O DINs have the most powerful modeling capability among
	IndexNet variants, but also introduce many extra parameters.
\end{remark}

M2O DINs have higher capacity than HINs and O2O DINs due to the use of standard convolution.
Another desirable property of IndexNet is that they
may be able to
predict the indices
from a large local feature map, e.g., $l(\mathcal{X}): \mathbb{R}^{2k\times 2k\times C}\rightarrow\mathbb{R}^{k\times k\times1}$.
An intuition behind this idea is that, if one identifies a local maximum point from a $k\times k$ region, its surrounding $2k\times2k$ region can further support whether this point is a part of a boundary or
only
an isolated noise point. This idea can be easily implemented by enlarging the convolutional kernel size and with appropriate padding.

In Table~\ref{tab:complexity_comparison}, we summarize the model complexity of different index networks used at a single downsampling and upsampling stage. We assume the convolution kernel has a size of $K\times K$ applied on a $C$-channel feature map. The number of parameters in BN layers is excluded. When considering weak context, we assume the kernel size is $2K\times 2K$. Since $C\gg K$, generally we have the model complexity \textit{M2O DINs}$>$\textit{HINs}$>$\textit{Unshared Stagewise O2O DINs}$>$\textit{Shared Stagewise O2O DINs}$>$\textit{Modelwise O2O DINs}.

\begin{table}[!t] \scriptsize
	\centering
	\addtolength{\tabcolsep}{-5pt}
	\renewcommand\arraystretch{1.0}
	\caption{A Comparison of Model Complexity of Different Index Networks}
	\label{tab:complexity_comparison}
	\vspace{-5pt}
	\begin{threeparttable}
		\begin{tabular}{l|c|c}
			\hline
			IndexNet & Type & \# Param.\\
			\hline
			\multirow{3}{*}{HINs} 							& L		& $K\times K\times C\times4$\\
															& NL	& $K\times K\times C\times2C+2C\times4$\\
															& NL+C  & $2K\times 2K\times C\times2C+2C\times4$\\
			\hline
			\multirow{3}{*}{Modelwise O2O DINs} 			& L		& $(K\times K)\times4$\\
															& NL	& $(K\times K\times2+2)\times4$\\
															& NL+C	& $(2K\times 2K\times2+2)\times4$\\
			\hline
			\multirow{3}{*}{Shared Stagewise O2O DINs} 		& L		& $(K\times K)\times4$\\
															& NL	& $(K\times K\times2+2)\times4$\\
															& NL+C	& $(2K\times 2K\times2+2)\times4$\\
			\hline
			\multirow{3}{*}{Unshared Stagewise O2O DINs} 	& L		& $(K\times K\times C)\times4$\\
															& NL	& $(K\times K\times 2C+2C\times C)\times4$\\
															& NL+C  & $(2K\times 2K\times 2C+2C\times C)\times4$\\
			\hline
			\multirow{3}{*}{M2O DINs} 						& L		& $(K\times K\times C\times C)\times4$\\
															& NL	& $(K\times K\times C\times 2C+2C\times C)\times4$\\
															& NL+C  & $(2K\times 2K\times C\times 2C+2C\times C)\times4$\\
			\hline

		\end{tabular}
		\begin{tablenotes}
			\scriptsize
			\item L: Linear; NL: Nonlinear; C: Context.
		\end{tablenotes}
	\end{threeparttable}
\end{table}

\section{Guided Upsampling or Blind Upsampling: A Reconstruction-Based Justification}

Here we introduce the concept of \textit{guided upsampling} and \textit{blind upsampling} to summarize existing upsampling operators.
Particularly, %
here we
present a %
comparison between two data-dependent upsampling operators---IndexNet and CARAFE~\cite{jiaqi2019carafe}. In addition, we
show results of an
image reconstruction task on synthetic data,  highlighting
the difference between guided upsampling and blind upsampling.

\subsection{Guided Upsampling vs.\  Blind Upsampling}

By blind upsampling, we mean that an upsampling operator that is pre-defined with fixed parameters.
Guided upsampling, instead,
is guided with the
involvement
of auxiliary information.
Thus,
most widely-used upsampling operators
peform
blind upsampling. These operators include nearest-neighbor (NN) interpolation, bilinear interpolation, space-to-depth and deconvolution. It is worth noting that the recent data-dependent upsampling operator CARAFE is also a blind upsampling operator.
By contrast, guided upsampling operators are rare in literature. Max unpooling,
albeit
being simple, is a guided upsampling operator. The auxiliary information used in upsampling comes from the max-pooling indices.
Thefore,
our proposed IndexNet clearly implements guided upsampling, with inferred dynamic indices as the auxiliary information.

A taxonomy of commonly-used upsampling operators and their corresponding downsampling operators is summarized in Table~\ref{tab:upsampling}.
An upsampling operator should generally have an 
corresponding
 downsampling operator, and vice versa.

\begin{table}[!t] \small
	\captionsetup{font=small,singlelinecheck=true}
	\centering
	\addtolength{\tabcolsep}{0pt}
	\renewcommand\arraystretch{1.0}
	\caption{Blind Upsampling and Guided Upsampling Operators}
	\label{tab:upsampling}
	\vspace{-5pt}
	\begin{tabular}{c|cc}
	    \hline
		 & Upsampling & Downsampling \\
		\hline
		\multirow{5}{*}{\parbox{2cm}{\centering Blind Upsampling}} & NN Interpolation & Average Pooling\\
		                                  & Bilinear Interpolation & Convolution\\
		                                  & Deconvolution & Convolution\\
		                                  & Space-to-Depth & Depth-to-Space\\
		                                  & CARAFE & Convolution\\
		\hline
		\multirow{2}{*}{\parbox{2cm}{\centering Guided Upsampling}} & Max Unpooling & Max Pooling\\
		                                   & Indexed Upsampling & Indexed Pooling\\
		\hline
	\end{tabular}
\end{table}

The main difference here is that
guided upsampling is made possible to exploit extra information to better recover the spatial information during upsampling.
Thus,
it is important that the spatial information is properly encoded during downsampling and is transferred to unsampling.

\subsection{IndexNet vs.\  CARAFE}
\label{ssec:comparison}

Both IndexNet and CARAFE are one of the %
few
attempts pursuing the idea of data-dependent upsampling.
The similarities include:
\begin{enumerate}[label=\roman*)]
	\item
	They both are
	related to
	dynamic networks.

	\item Both are parametric upsampling operators;
	\item CARAFE and HINs both
	perform
	holistic upsampling.
	\item CARAFE also learns
	an
	index function. The index function has an identical form to Eq.~\eqref{eq:dconv}, but with dynamic and normalized $\mat W$. In this sense, CARAFE may be considered as a single-input version of $\mathcal{IU}$, where the index map is generated internally.
\end{enumerate}
The differences are:
\begin{enumerate}[label=\roman*)]
	\item
	CARAFE is a blind upsampling operator, while IndexNet implements guided upsampling;
    \item The reassembly kernels in CARAFE are generated conditioned on the low-resolution decoder feature map. The index maps predicted by IndexNet, however, build upon the high-resolution encoder feature map, before spatial information is lost;
    \item In IndexNet, each upsampled feature point only associates with a single point in the low-resolution feature map. From low resolution to high resolution, it is a one-to-many mapping. In CARAFE, each upsampled point is a weighted sum of a local region from the low-resolution feature map. This is a many-to-one mapping;

    \item Compared to CARAFE which is presented as a single upsampling operator, IndexNet is a
    more general  framework.
\end{enumerate}
In particular, the key difference lies in the intermediate path that allows spatial information to be visible to upsampling. To further demonstrate the benefit of this intermediate path, we present an
image reconstruction experiment on synthetic data, namely, the
Fashion-MNIST dataset~\cite{xiao2017fashion}.

\subsection{Fashion-MNIST Image Reconstruction}

The idea is that, if an upsampling operator can recover spatial information well from downsampled feature maps, the reconstructed output should be visually
closer
to the input image. The quality of reconstruction results can be a good indicator how well spatial information is recovered by an upsampling operator.

\textit{Network Architecture and Baselines.} We use a standard encoder-decoder architecture. Let \textbf{C}($k$) denote a 2D convolutional layer with $k$-channel $3\times3$ filters, followed by BN and ReLU. \textbf{$D$} represents a downsampling operator with a downsampling ratio of $2$, and \textbf{$U$} an upsampling operator with an upsampling ratio of $2$. The reconstruction network can therefore be defined by \textbf{C}($32$)-\textbf{$D$}-\textbf{C}($64$)-\textbf{$D$}-\textbf{C}($128$)-\textbf{$D$}-\textbf{C}($256$)-\textbf{C}($128$)-\textbf{$U$}-\textbf{C}($64$)-\textbf{$U$}-\textbf{C}($32$)-\textbf{$U$}-\textbf{C}($1$). Note that BN and ReLU are not included in the last \textbf{C}($1$).
We build the following baselines:
\begin{enumerate}[label=\roman*)]
    \item Average Pooling--NN interpolation (AvgPool--NN);
    \item stride-2  Convolution--Bilinear interpolation (Conv\textsubscript{/2}--Bilinear);
    \item Space-to-Depth--Depth-to-Space (S2D--D2S);
    \item stride-2 Convolution--2-stride Deconvolution (Conv\textsubscript{/2}--Deconv\textsubscript{/2});
    \item stride-2  Convolution--CARAFE (Conv\textsubscript{/2}--CARAFE);
    \item Max Pooling--Max Unpooling (MaxPool--MaxUnpool);
    \item Indexed Pooling--Indexed Upsampling ($\mathcal{IP}$--$\mathcal{IU}$).
\end{enumerate}

\begin{table}[!t]
    \captionsetup{font=small,singlelinecheck=true}
	\centering
	\addtolength{\tabcolsep}{2pt}
	\renewcommand\arraystretch{1.0}
	\caption{Performance of Image Reconstruction on the Fashion-MNIST Dataset}
	\label{tab:reconstruction}
	\vspace{-5pt}
    \begin{tabular}{l|cccc}
        \hline
         & PSNR & SSIM & MAE & MSE \\
         \hline
        AvgPool--NN                                                 & 25.88 & 0.9811 & 0.0259 & 0.0509\\
        Conv\textsubscript{/2}--Bilinear                            & 24.45 & 0.9726 & 0.0320 & 0.0600\\
        S2D--D2S                                                    & 28.93 & 0.9901 & 0.0204 & 0.0358\\
        Conv\textsubscript{/2}--Deconv\textsubscript{/2}            & 28.75 & 0.9903 & 0.0187 & 0.0366\\
        Conv\textsubscript{/2}--CARAFE                              & 25.55 & 0.9798 & 0.0277 & 0.0529\\
        MaxPool--MaxUnpool                                          & 29.33 & 0.9920 & 0.0202 & 0.0342\\
        $\mathcal{IP}$--$\mathcal{IU}$\textsuperscript{*}           & 37.83 & 0.9989 & 0.0089 & 0.0128\\
        $\mathcal{IP}$--$\mathcal{IU}$\textsuperscript{$\dagger$}   & 45.93 & 0.9998 & 0.0032 & 0.0051\\
        $\mathcal{IP}$--$\mathcal{IU}$\textsuperscript{$\ddagger$}  & \textbf{48.37} & \textbf{0.9999} & \textbf{0.0026} & \textbf{0.0038}\\
        \hline
    \end{tabular}

    \begin{tablenotes}
    \item[1] $^*$ denotes Modelwise O2O DIN; $^\dagger$ indicates HIN; $^\ddagger$ refers to M2O DIN. All IndexNets are with nonlinearity and weak context. The best performance is boldfaced.
    \end{tablenotes}
\end{table}

\textit{Training Details.} The Fashion-MNIST dataset~\cite{xiao2017fashion} includes $60,000$ training images and $10,000$ testing images. The input images are resized to $32\times32$. $\ell_1$ loss is used in training. The initial learning rate is set to $0.01$. We train the network for $100$ epochs with a batch size of $100$. The learning rate is decreased by $\times10$ at the $50$-th, $70$-th and $85$-th epoch, respectively. We report Peak Signal-to-Noise Ratio (PSNR), Structural SIMilarity index (SSIM), Mean Absolute Error (MAE) and root Mean Square Error (MSE).

\begin{figure*}[!t]
    \centering
    \includegraphics[width=.6999\linewidth,angle=0]{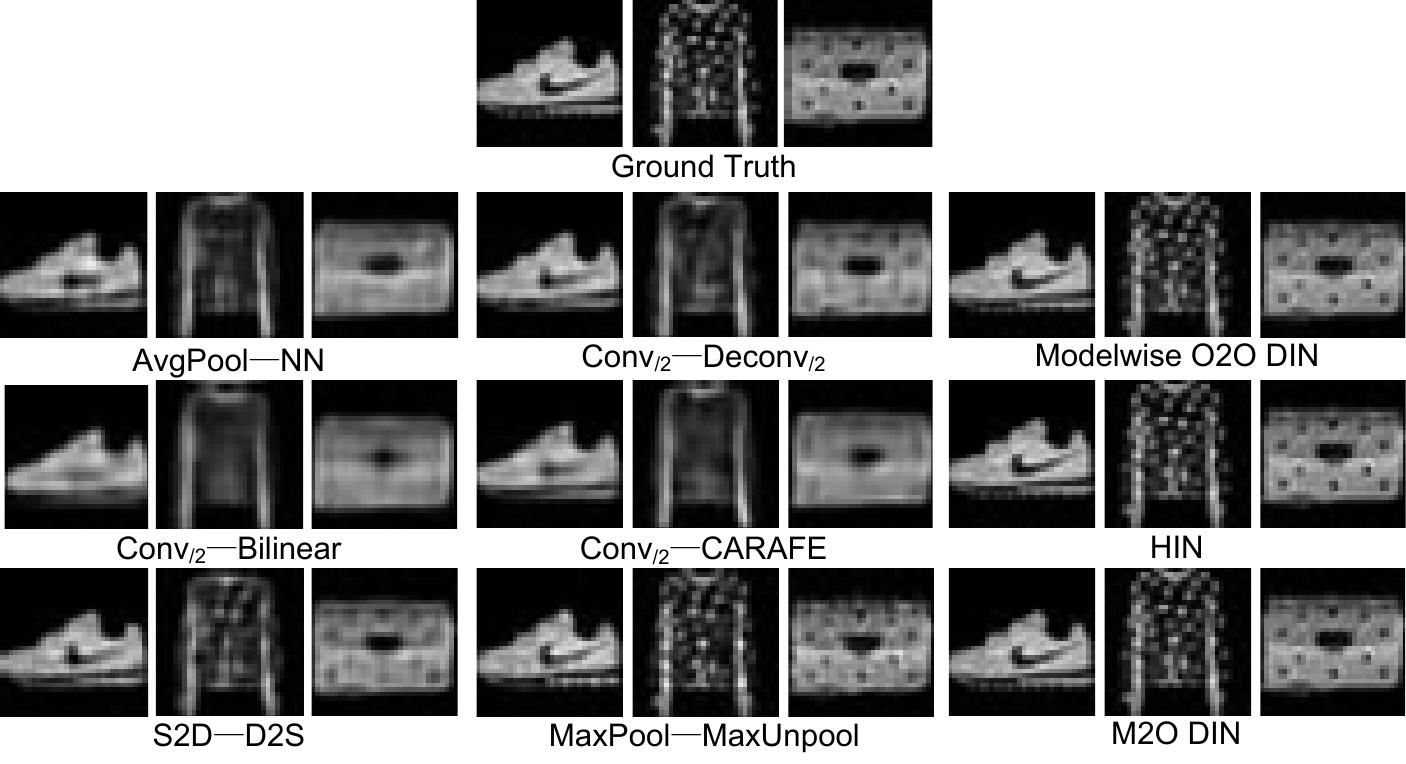}\vspace{-10pt}
    \caption{Image reconstruction results on the Fashion-MNIST dataset.}
    \label{fig:recover}
\end{figure*}

\textit{Discussions.} Quantitative and qualitative results are shown in Table~\ref{tab:reconstruction} and Fig.~\ref{fig:recover}, respectively. We observe in Fig.~\ref{fig:recover} that, all baselines that exploit blind upsampling fail to reconstruct the input images. While in some cases they may produce reasonable reconstructions, images of trousers and high-heeled shoes for instance, in most cases these baselines generate blurred results when complex textural patterns appear, tops and wallets for example. By contrast, other baselines that leverage guided upsampling produce visually pleasing reconstructions, in all circumstances. In particular, by comparing MaxPool--MaxUnpool with $\mathcal{IP}$--$\mathcal{IU}$, the former tends to yield jittering artifacts. This suggests index maps extract and
encode
richer spatial information than max-pooling indices. In addition, by disabling the intermediate path, $\mathcal{IP}$--Bilinear leads to significantly poor reconstructions, which means that
 the intermediate path  matters. The differences between two upsampling paradigms are also supported by the numerical results in Table~\ref{tab:reconstruction} where guided upsampling exhibits significantly better PSNR and lower errors than blind upsampling. Indeed, the intermediate path distinguishes guided upsampling from blind upsampling, and also IndexNet from CARAFE.

\section{Applications}

In this section, we show several applications of IndexNet on the tasks of image matting, image denoising, semantic segmentation,
and
monocular depth estimation.

\subsection{Image Matting}

We first evaluate IndexNet on the task of image matting. Image matting is defined as a problem of estimating soft foreground from images. This problem is ill-posed due to the fact that solving a linear system of $7$ unknown variables with only $3$ known inputs: given the RGB color at pixel $i$, $I_i$, one needs to estimate the corresponding foreground color $F_i$, background color $B_i$, and matte $\alpha_i$, such that $I_i=\alpha_iF_i+(1-\alpha_i)B_i$, for any $\alpha_i\in[0,1]$. Previous methods have extensively studied this problem from a low-level view~\cite{chen2013knn,chuang2001bayesian,he2011global,levin2008closed}; and particularly, they have been designed to solve the above matting equation. Despite being theoretically elegant, these methods heavily rely on the color cues, rendering failures of matting in general natural scenes where colors are not reliable. With the tremendous success of deep CNNs in high-level vision tasks~\cite{girshick2014rich,krizhevsky2012imagenet,long2015fully}, deep matting methods are emerging.
Recently %
deep image matting was proposed~\cite{xu2017deep}. In \cite{xu2017deep} the authors presented the first deep image matting approach (DeepMatting) based on SegNet~\cite{badrinarayanan2017segnet} and significantly outperformed other competitors. In this application, we
use
DeepMatting as our baseline.
\begin{figure}[!tb]
	\captionsetup{font=small,singlelinecheck=true}
	\setlength{\abovecaptionskip}{10pt}
	\centering
	\includegraphics[width=2.8in,angle=0]{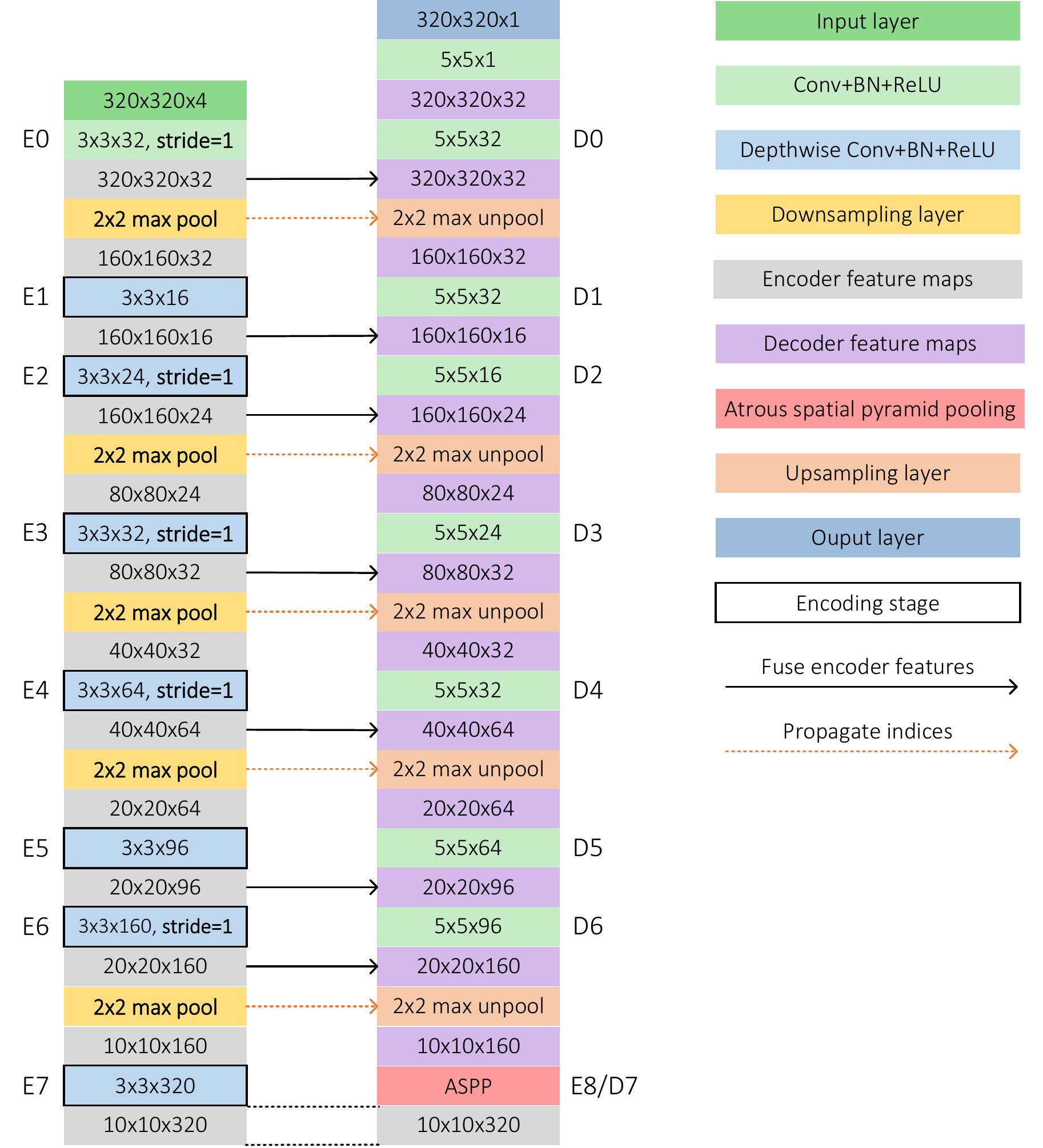}\vspace{-10pt}
	\caption{Customized MobileNetv2-based encoder-decoder network architecture. Our modifications are boldfaced.}
	\label{fig:network_architecture}
\end{figure}
Image matting is particularly suitable for evaluating the effectiveness of IndexNet, because the quality of learned indices can be visually observed from inferred alpha mattes. We conduct experiments on the Adobe Image Matting dataset~\cite{xu2017deep}. This is so far the largest publicly available matting dataset. The training set has $431$ foreground objects and ground-truth alpha mattes.
Each foreground is composited with $100$ background images randomly chosen from MS COCO~\cite{lin2014microsoft}. The validation set termed Composition-1k includes $100$ unique objects. Each of them is composited with 10 background images chosen from Pascal VOC~\cite{everingham2010pascal}. Overall, we have $43,100$ training images and $1,000$ testing images. We evaluate the results using widely-used Sum of Absolute Differences (SAD), root Mean Square Error (MSE), and perceptually-motivated Gradient (Grad) and Connectivity (Conn) errors~\cite{rhemann2009perceptually}. The evaluation code implemented by~\cite{xu2017deep} is used. In what follows, we first describe our modified MobileNetv2-based architecture and training details. We then perform extensive ablation studies to justify choices of model design, make comparisons of different index networks, and visualize learned indices.

\subsubsection{Network Architecture and Implementation Details}

Here we describe the network architecture %
and
training details.
\begin{table*}[!t]
	\centering
	\renewcommand\arraystretch{1.0}
	\addtolength{\tabcolsep}{1.5pt}
	\caption{Ablation Study of Design Choices}
	\label{tab:architecture}
	\vspace{-5pt}
	\begin{threeparttable}
		\begin{tabular}{llccccc|ccccc}
			\hline
			No. & Architecture					& Backbone		& Fusion 	& Indices 	& Context 	& OS 	& SAD 	& MSE 	& Grad & Conn \\
			\hline
			B1	& DeepLabv3+~\cite{chen18v3}	& MobileNetv2	& Concat	& No		& ASPP		& 16	& 60.0	& 0.020	& 39.9 & 61.3 \\
			B2 	& RefineNet~\cite{lin2017refine}& MobileNetv2	& Skip 		& No 		& CRP 		& 32 	& 60.2	& 0.020	& 41.6 & 61.4 \\
			\rowcolor{mygray}
			B3 	& SegNet~\cite{xu2017deep}		& VGG16			& No		& Yes		& No		& 32	& \textbf{54.6} 	& \textbf{0.017} & 36.7 & 55.3 \\
			\rowcolor{mygray}
			B4 	& SegNet						& VGG16			& No		& No		& No		& 32	& 122.4 & 0.100 & 161.2 & 130.1 \\
			\rowcolor{mygray}
			B5 	& SegNet		 				& MobileNetv2	& No 		& Yes 		& No 		& 32	& 60.7	& 0.021	& 40.0 & 61.9 \\
			\rowcolor{mygray}
			B6 	& SegNet		 				& MobileNetv2	& No 		& No 		& No 		& 32	& 78.6  & 0.031 & 101.6& 82.5 \\
			B7 	& SegNet		 				& MobileNetv2	& No 		& Yes 		& ASPP 		& 32	& 58.0	& 0.021	& 39.0 & 59.5 \\
			B8 	& SegNet		 				& MobileNetv2	& Skip 		& Yes 		& No 		& 32	& 57.1 	& 0.019	& 36.7 & 57.0 \\
			B9 	& SegNet		 				& MobileNetv2	& Skip 		& Yes 		& ASPP 		& 32	& 56.0	& \textbf{0.017}	& 38.9 & 55.9 \\
			B10 & UNet	 						& MobileNetv2	& Concat 	& Yes 		& No 		& 32 	& 54.7	& \textbf{0.017}	& 34.3 & \textbf{54.7} \\
			B11 & UNet	 						& MobileNetv2	& Concat 	& Yes 		& ASPP 		& 32	& 54.9	& \textbf{0.017} & \textbf{33.8} & 55.2 \\
			\hline
		\end{tabular}
		\begin{tablenotes}
			\scriptsize
			\item Fusion: fuse encoder features; Indices: max-pooling indices (where Indices is `No', bilinear interpolation is used for upsampling); CRP: chained residual pooling~\cite{lin2017refine}; ASPP: atrous spatial pyramid pooling~\cite{chen18v3}; OS: output stride. The lowest errors are boldfaced.
		\end{tablenotes}
	\end{threeparttable}
\end{table*}

\textit{Network Architecture}. We build our model based on MobileNetv2~\cite{sandler2018mobilenetv2} with only slight modifications to the backbone.
We choose MobileNetv2
for its
lightweight model and fast inference.
The basic network configuration is shown in Fig.~\ref{fig:network_architecture}. It also follows the encoder-decoder paradigm same as SegNet. We simply change all 2-stride convolution to be 1-stride and attach 2-stride $2\times2$ max pooling after each encoding stage for downsampling, which allows us to extract indices. If applying the IndexNet idea, max pooling and unpooling layers can be replaced with $\mathcal{IP}$ and $\mathcal{IU}$, respectively. We also investigate alternative ways for low-level feature fusion and whether encoding context (Section~\ref{subsubsec:ablation_study}). Note that, the matting refinement stage~\cite{xu2017deep} is not
applied here.

\textit{Training Details}. To enable a direct comparison with deep matting~\cite{xu2017deep}, we follow the same training configurations used in~\cite{xu2017deep}. The 4-channel input concatenates the RGB image and its trimap. We follow exactly the same data augmentation strategies, including $320\times320$ random cropping, random flipping, random scaling, and random trimap dilation.
We use a combination of the alpha prediction loss and the composition loss during training as in~\cite{xu2017deep}. Only losses from the unknown region of the trimap are calculated. Encoder parameters are pretrained on ImageNet~\cite{deng2009imagenet}.
The parameters of the $4$-th input channel are initialized with zeros.
The Adam optimizer~\cite{kingma2015adam} is used. We update parameters with $30$ epochs (around $90,000$ iterations). The learning rate is initially set to $0.01$ and reduced by $10\times$ at the $20$-th and $26$-th epoch respectively. We use a batch size of $16$ and fix the BN layers of the backbone.

\begin{table}[!t]
	\centering
	\addtolength{\tabcolsep}{-0.75pt}
	\renewcommand\arraystretch{1.0}
	\caption{Results on the Composition-1k Testing Set}
	\label{tab:index_function}
	\vspace{-5pt}
	\begin{threeparttable}
		\begin{tabular}{>{\centering}p{.5cm}|>{\centering}p{.5cm}|c|c|c|c|c|c}
			\hline
			\multicolumn{2}{c|}{Method} 				& \#Param. & GFLOPs & SAD & MSE & Grad & Conn\\
			\hline
			\multicolumn{2}{c|}{B3~\cite{xu2017deep}} 	& 130.55M & 32.34 & 54.6 & 0.017 & 36.7 & 55.3 \\
			\multicolumn{2}{c|}{B11} 					& 3.75M	  & 4.08  & 54.9 & 0.017 & 33.8 & 55.2 \\
			\multicolumn{2}{c|}{B11-1.4} 				& 8.86M	  & 7.61  & 55.6 & 0.016 & 36.4 & 55.7 \\
			\multicolumn{2}{c|}{B11-carafe} 			& 4.06M	  & 5.01 & 50.2 & 0.015 & 27.9 & 50.0\\
			\multicolumn{2}{c|}{HMI} 					& 3.75M   & 4.08  & 56.5 & 0.021 & 33.0 & 56.4 \\
			\hline
			NL & C & $\Delta$ \\
			\hline
			& & \multicolumn{6}{c}{HINs}\\
			\hline
			& 									 		& +4.99K  & 4.09  & 55.1 & 0.018 & 32.1 & 55.2\\
			\checkmark & 						 		& +0.26M  & 4.22  & 50.6 & 0.015 & 27.9 & 49.4\\
			\checkmark & \checkmark 			 		& +1.04M  & 4.61  & 49.5 & 0.015 & \textbf{25.6} & 49.2\\
			\hline
			& & \multicolumn{6}{c}{Modelwise O2O DINs}\\
			\hline
			& 							 		 		& +16     & 4.08  & 57.3 & 0.017 & 37.3 & 57.4\\
			\checkmark & 						 		& +56 	  & 4.08  & 52.4 & 0.016 & 30.1 & 52.2\\
			\checkmark & \checkmark 			 		& +152    & 4.08  & 59.1 & 0.018 & 39.0 & 59.7\\
			\hline
			& & \multicolumn{6}{c}{Shared Stagewise O2O DINs}\\
			\hline
			&							 		 		& +80  	  & 4.08  & 48.9 & 0.014 & 26.2 & 48.0\\
			\checkmark & 						 		& +280    & 4.08  & 51.1 & 0.016 & 30.2 & 50.7\\
			\checkmark & \checkmark 			 		& +760    & 4.08  & 56.0 & 0.016 & 37.5 & 55.9\\
			\hline
			& & \multicolumn{6}{c}{Unshared Stagewise O2O DINs}\\
			\hline
			&          					                & +4.99K  & 4.09  & 50.3 & 0.015 & 33.7 & 50.0\\
			\checkmark &            	                & +17.47K & 4.10  & 50.6 & 0.016 & 26.5 & 50.3\\
			\checkmark & \checkmark 	                & +47.42K & 4.15  & 50.2 & 0.016 & 26.8 & 49.3\\
			\hline
			& & \multicolumn{6}{c}{M2O DINs}\\
			\hline
			&          				             		& +0.52M & 4.34   & 51.0 & 0.015 & 33.7 & 50.5\\
			\checkmark &                         		& +1.30M & 4.73   & 48.9 & 0.015 & 32.1 & 47.9\\
			\checkmark & \checkmark              		& +4.40M & 6.30   & \textbf{45.8} & \textbf{0.013} & 25.9 & \textbf{43.7}\\
			\hline
			\multicolumn{4}{c|}{DeepMatting w. Refinement~\cite{xu2017deep}} & 50.4 & 0.014 & 31.0 & 50.8 \\
			\hline
		\end{tabular}
		\begin{tablenotes}
			\scriptsize
			\item NL: Non-Linearity; C: Context. $\Delta$ indicates increased parameters compared to B11. GFLOPs are measured on a $224\times224\times4$ input. The lowest errors are boldfaced.
		\end{tablenotes}
	\end{threeparttable}
\end{table}

\begin{figure*}[!tb]
	\setlength{\abovecaptionskip}{10pt}
	\centering
	\includegraphics[width=\linewidth,angle=0]{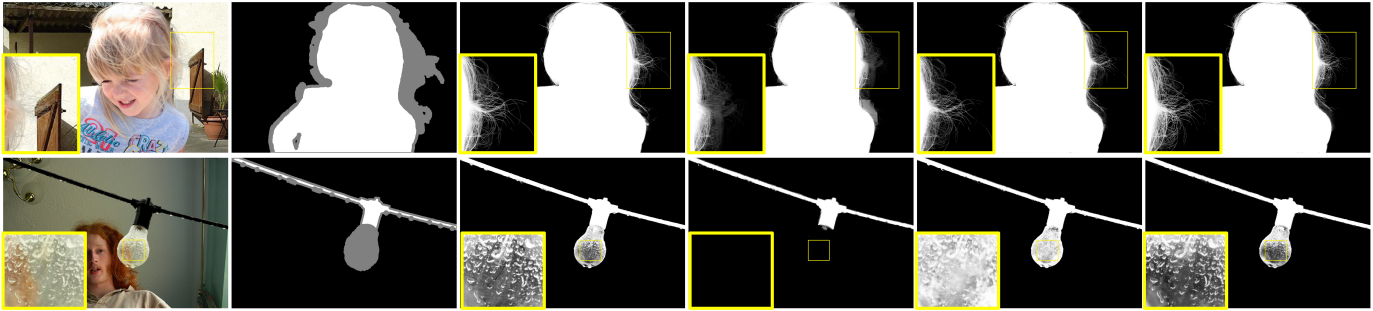}\vspace{-10pt}
	\caption{Qualitative results on the Composition-1k testing set. From left to right, the original image, trimap, ground-truth alpha matte, Closed-form Matting~\cite{levin2008closed}, DeepMatting~\cite{levin2008closed}, and ours (M2O DIN with `Nonlinearity+Context').}
	\label{fig:composition-1k-results}
\end{figure*}

\subsubsection{Results on the Adobe Image Matting Dataset}
\label{subsubsec:ablation_study}

\textit{Ablation Study on Model Design}. To establish a better baseline comparable to DeepMatting, here we first investigate strategies for fusing low-level features (no fusion, skip fusion as in ResNet~\cite{he2016deep} or concatenation as in UNet~\cite{ronneberger2015u}) and whether encoding context for image matting. $11$ baselines are consequently built to justify model design. Results on the Composition-1k testing set are reported in Table~\ref{tab:architecture}. B3 is cited from~\cite{xu2017deep}. We can make the following observations:
\begin{enumerate}[label=\roman*)]
	\item Indices are of great importance. Matting can significantly benefit from only indices (B3 vs.\ B4, B5 vs.\ B6);
	\item State-of-the-art semantic segmentation models cannot be directly applied to image matting (B1/B2 vs.\ B3);
	\item Fusing low-level features help, and concatenation is better than skip connection but at a cost of increased computation (B6 vs.\ B8 vs.\ B10 or B7 vs.\ B9 vs.\ B11);
    \item
    Modules such as ASPP may improve the results (e.g., B6 vs.\ B7 or B8).

	\item A MobileNetv2-based matting model can work as well as a VGG-16-based one (B3 vs.\ B11).
\end{enumerate}
For the following experiments, we now mainly use B11.

\textit{Ablation Study on Index Networks}. Here we compare different index networks and justify their effectiveness. The configurations of index networks used in the experiments follow Figs.~\ref{fig:holistic_networks} and~\ref{fig:depthwise_networks}. We primarily investigate the $2\times2$ kernel with a stride of $2$. Whenever the weak context is considered, we use a $4\times4$ kernel in the first convolutional layer of index networks. To highlight the effectiveness of HINs, we further build a baseline called \textit{holistic max index} (HMI) where max-pooling indices are extracted from a squeezed feature map $\mathcal{X}'\in\mathbb{R}^{H\times W\times1}$. $\mathcal{X}'$ is generated by applying the max function along the channel dimension of $\mathcal{X}\in\mathbb{R}^{H\times W\times C}$. Furthermore, since IndexNet increases extra parameters, we introduce another baseline \textit{B11-1.4} where the width multiplier of MobilieNetV2 is adjusted to be $1.4$ to increase the model capacity. In addition, to compare IndexNet against CARAFE in this task, we build an additional baseline \textit{B11-carafe} where the unpooling operator in B11 is replaced with CARAFE. Results on the \mbox{Composition-1k} testing dataset are listed in Table~\ref{tab:index_function}. We observe that, most index networks reduce the errors notably, except for some low-capacity IndexNet modules (due to limited modeling capabilities). In particular, nonlinearity and the context generally have a positive effect on deep image matting, but they do not work effectively in O2O DINs. A possible reason may be that the limited dimensionality of the intermediate feature map is not sufficient to model complex patterns in matting. Compared to holistic max index, the direct baseline of HINs, the best HIN (``Nonlinearity+Context'') has at least $12.3\%$ relative improvement. Compared to B11, the baseline of DINs, M2O DIN with ``Nonlinearity+Context'' exhibits at least $16.5\%$ relative improvement. Notice that, our best model outperforms the DeepMatting approach~\cite{xu2017deep} that even has the refinement stage. In addition, according to the results of B11-1.4, the performance improvement does not come from increased parameters. Moreover, CARAFE also enhances matting performance, but it falls behind M2O DIN. Some qualitative results are shown in Fig.~\ref{fig:composition-1k-results}. Our predicted mattes show improved delineation for edges and textures like hair and water drops.

\begin{table}[!t] \small
	\captionsetup{font=small,singlelinecheck=true}
	\centering
	\addtolength{\tabcolsep}{-2pt}
	\renewcommand\arraystretch{1.0}
	\caption{Ablation Study of Different Normalization Choices on Index Maps}
	\label{tab:normalize}
	\vspace{-5pt}
	\begin{threeparttable}
		\begin{tabular}{cc|cccc}
			\hline
			Encoder & Decoder & SAD & MSE & Grad & Conn \\
			\hline
			sigmoid & sigmoid & 52.7 & 0.016 & 29.3 & 52.4 \\

			softmax & softmax & 51.6 & 0.015 & 29.2 & 51.6 \\

			softmax+sigmoid & softmax & 57.3 & 0.016 & 43.5 & 57.3 \\

			sigmoid+softmax & sigmoid & \textbf{45.8} & \textbf{0.013} & \textbf{25.9} & \textbf{43.7} \\
			\hline
		\end{tabular}
		\begin{tablenotes}
			\scriptsize
			\item The lowest errors are boldfaced.
		\end{tablenotes}
	\end{threeparttable}
\end{table}

\textit{Ablation Study on Index Normalization}. Index normalization
is important
for
the final performance. Here we justify this by evaluating different normalization choices.
Apart
from the sigmoid function used for the decoder and the sigmoid+softmax function for the encoder, we compare other three different combinations of normalization strategies listed in Table~\ref{tab:normalize}. The experiment is conducted based on M2O DIN with ``Nonlinearity+Context''.
It is clear that keeping the magnitude consistency during downsampling matters. In fact, both max pooling and average pooling satisfy this property naturally, and our normalization design is inspired from this fact.

\begin{figure}[!tb]
	\captionsetup{font=small,singlelinecheck=true}
	\setlength{\abovecaptionskip}{10pt}
	\centering
	\includegraphics[width=.46\textwidth]{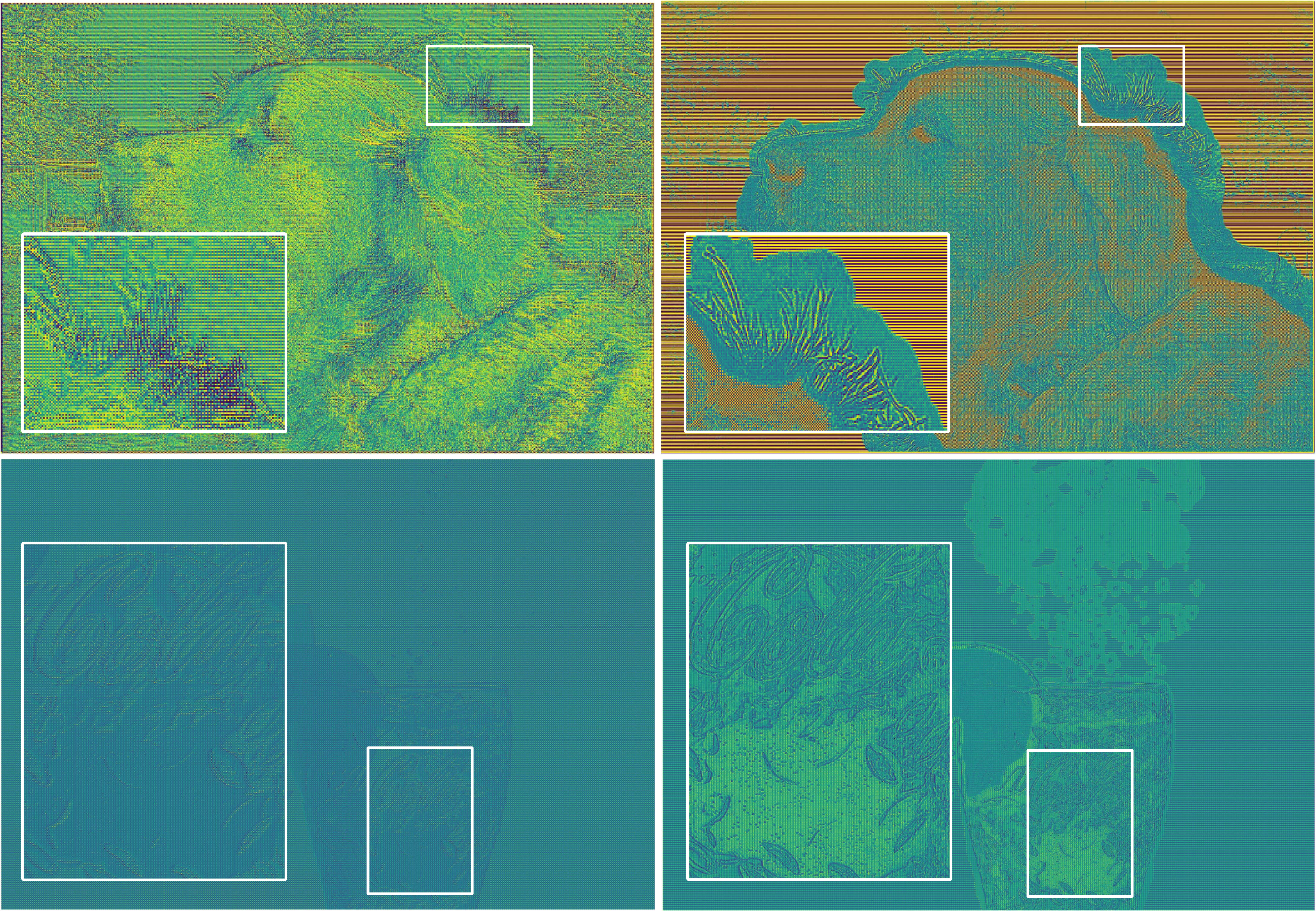}\vspace{-10pt}
	\caption{Visualization of the randomly initialized index map (left) and the learned index map (right) of HINs (top) and DINs (bottom). Best viewed
	on
	screen.}
	\label{fig:indices_visualization}
\end{figure}

\textit{Index Map Visualization}. It is interesting to see what indices are learned by IndexNet. For the holistic index, the index map itself is a 2D matrix and can be easily visualized. Regarding the depthwise index, we squeeze the index map along the channel dimension and calculate the average responses. Two examples of learned index maps are visualized in Fig.~\ref{fig:indices_visualization}. We observe that, initial random indices have poor delineation for edges, while learned indices automatically capture the complex structural and textual patterns, e.g., the fur of the dog, and even air bubbles in the water.

\subsection{Image Denoising}

The goal of image denoising is to recover a clean image $\mat x$ from a corrupted observation $\mat y$ following an image degradation model $\mat y=\mat x + \mat v$, where $\mat v$ is commonly assumed to be additive white Gaussian noise (AWGN) parameterized by $\sigma$. While such an assumption has been challenged in recent real-image denoising~\cite{chen2018learning,jaroensri2019generating}, we still follow the AWGN paradigm in evaluation because our focus is not to improve image denoising.
When deep CNNs are widely accepted, the data-driven paradigm
now
becomes the first-class choice for image denoising~\cite{mao2016image,zhang2017beyond}. Most deep denoising models are designed with the same high-level idea---processing the feature map without decreasing its spatial resolution. Indeed, it has been observed that, when the feature map is downsampled, the performance drops remarkably~\cite{mao2016image}.
For such networks, although the model parameters largely reduce,
computational complexity of
training and inference becomes much heavier.

We show that, by inserting IndexNet into a denoising model, it can effectively compensate the loss of spatial information, achieving performance comparable to or even better than the network without downsampling.
Thus,
despite the number of parameters increases,
computation us much reduced.
We choose DnCNN~\cite{zhang2017beyond} as our baseline
to demonstrate this
on standard benchmarks. We follow the experimental setting of~\cite{chen2016trainable} that uses a 400-image training set. The performance is reported on a 68-image Berkeley segmentation dataset (BSD68) and the other 12-image test set (Set12). The networks are trained for Gaussian denoising, with three noise levels, i.e., $\sigma=15,25$ and $50$. PSNR and SSIM are used as evaluation metrics.

\begin{figure}[!tb]
	\setlength{\abovecaptionskip}{10pt}
	\centering
	\includegraphics[width=\linewidth,angle=0]{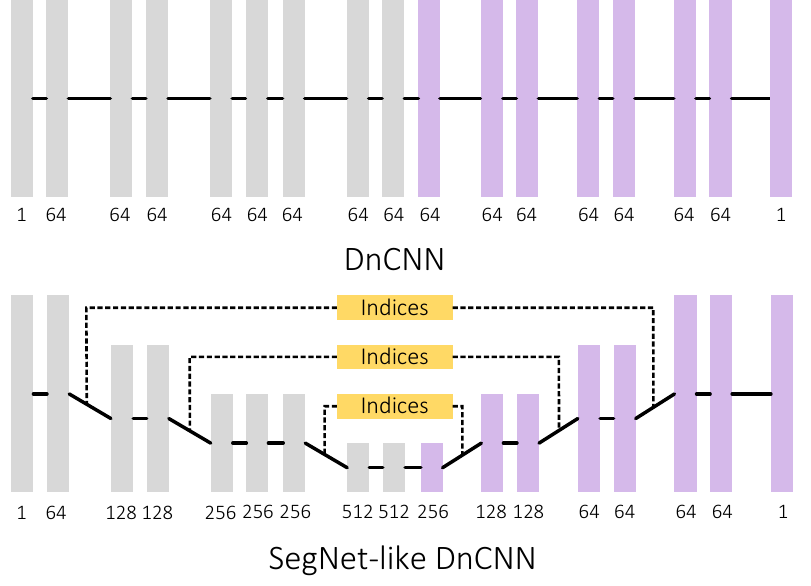}\vspace{-10pt}
	\caption{The DnCNN architecture and our modified SegNet-like DnCNN.}
	\label{fig:dncnn}
\end{figure}

\begin{table*}[!t]
	\captionsetup{font=small,singlelinecheck=true}
	\centering
	\renewcommand\arraystretch{1.0}
	\caption{Average PSNR (dB) and SSIM Results of Various Noise Levels on the BSD68 and Set12 Image Denoising Benchmarks}
	\label{tab:indexnet_denoising}
	\vspace{-5pt}
	\begin{threeparttable}
		\begin{tabular}{>{\centering}p{.85cm}|>{\centering}p{1.25cm}|c|c|ccc|ccc}
			\hline
			\multicolumn{2}{c|}{Method} & \#Param. & GFLOPs & \multicolumn{3}{c|}{BSD68} & \multicolumn{3}{c}{Set12}\\
			\hline
			\multicolumn{2}{c|}{Noise Level} & & & 15 & 25 & 50 & 15 & 25 & 50\\
			\hline
			\multicolumn{2}{c|}{DnCNN~\cite{zhang2017beyond}} 	& 0.56M & 25.89 & 31.74/0.9410 & 29.22/0.9015 & 26.23/0.8269 & 32.87/0.9544 & 30.42/0.9296 & 27.17/0.8775 \\
			\multicolumn{2}{c|}{DnCNN-SegNet}					& 7.09M & 18.14 & 30.74/0.9278 & 28.27/0.8752 & 24.88/0.7437 & 31.91/0.9395 & 28.98/0.8881 & 24.99/0.7485 \\
			\multicolumn{2}{c|}{DnCNN-max-carafe}				    & 7.29M & 19.11 & 25.70/0.7578 & 21.41/0.5670 & 15.40/0.2988 & 24.64/0.6997 & 20.19/0.4965 & 14.22/0.2489\\
			\multicolumn{2}{c|}{DnCNN-avg-carafe}				    & 7.29M & 19.11 & 25.70/0.7578 & 21.43/0.5673 & 15.56/0.2968 & 24.64/0.6997 & 20.20/0.4968 & 14.26/0.2460\\
			\multicolumn{2}{c|}{DnCNN-conv-carafe}				    & 7.29M & 15.23 & 25.70/0.7578 & 21.43/0.5672 & 15.50/0.2994 & 24.64/0.6997 & 20.20/0.4967 & 14.22/0.2496\\
			\hline
			NL & C  & $\Delta$\\
			\hline
			& & & & \multicolumn{6}{c}{HINs}\\
			\hline
			& 	 									& +7.17K & 18.16 & 31.13/0.9357 & 29.02/0.8997 & 26.29/0.8281 & 32.71/0.9536 & 30.28/0.9285 & 27.20/0.8789 \\
			\checkmark &  							& +0.69M & 19.30 & 31.15/0.9356 & 29.01/0.8999 & 26.29/0.8301 & 32.77/0.9537 & 30.36/0.9295 & 27.18/0.8799 \\
			\checkmark & \checkmark  				& +2.76M & 22.75 & 31.20/0.9365 & 29.05/0.9004 & 26.30/0.8305 & 32.79/0.9541 & 30.37/0.9300 & 27.22/0.8804 \\
			\hline
			& & & & \multicolumn{6}{c}{Modelwise O2O DINs}\\
			\hline
			& 										& +16  & 18.14 & 31.22/0.9366 & 29.06/0.9002 & 25.84/0.8294 & 32.83/0.9545 & 30.42/0.9302 & 26.21/0.8782 \\
			\checkmark & 							& +56  & 18.14 & 30.64/0.9255 & 27.39/0.8391 & 24.15/0.6776 & 31.92/0.9386 & 27.97/0.8330 & 24.14/0.6747 \\
			\checkmark & \checkmark 				& +152 & 18.14 & 30.87/0.9296 & 27.70/0.8617 & 24.09/0.6939 & 32.23/0.9432 & 28.31/0.8677 & 23.85/0.6634 \\
			\hline
			& & & & \multicolumn{6}{c}{Shared Stagewise O2O DINs}\\
			\hline
			& 										& +48  & 18.14 & 31.14/0.9364 & 29.05/0.9002 & 26.32/0.8310 & 32.80/0.9542 & 30.41/0.9302 & 27.24/0.8807 \\
			\checkmark & 	 						& +168 & 18.14 & 31.20/0.9365 & 28.97/0.9000 & 26.18/0.8272 & 32.83/0.9545 & 30.43/0.9302 & 27.24/0.8801 \\
			\checkmark & \checkmark 				& +456 & 18.14 & 31.22/0.9366 & 29.07/0.9004 & 26.31/0.8311 & 32.82/0.9543 & 30.41/0.9300 & 27.27/0.8814 \\
			\hline
			& & & & \multicolumn{6}{c}{Unshared Stagewise O2O DINs}\\
			\hline
			& 	 									& +7.17K & 18.16 & 31.17/0.9366 & 28.25/0.8944 & 25.02/0.8235 & 32.80/0.9544 & 30.23/0.9286 & 26.41/0.8675 \\
			\checkmark & 							& +25.1K & 18.19 & 31.25/0.9368 & 29.06/0.9002 & 26.33/0.8306 & 32.77/0.9541 & 30.43/0.9303 & 27.29/0.8814 \\
			\checkmark & \checkmark 	 			& +68.1K & 18.32 & 31.21/0.9364 & 27.68/0.8740 & 26.33/0.8312 & 32.83/0.9544 & 30.32/0.9288 & 27.24/0.8807 \\
			\hline
			& & & & \multicolumn{6}{c}{M2O DINs}\\
			\hline
			& 	 									& +1.38M & 20.44 & 31.22/0.9365 & 29.03/0.9005 & 26.33/0.8316 & 32.82/0.9544 & 30.42/0.9302 & 27.28/0.8812 \\
			\checkmark &  							& +3.45M & 23.88 & 31.23/0.9368 & 29.07/0.9002 & 26.26/0.8278 & 32.84/0.9546 & 30.44/0.9304 & 27.28/0.8808 \\
			\checkmark & \checkmark  				& +11.7M & 37.67 & 31.23/0.9365 & 29.06/0.8996 & 26.34/0.8315 & 32.82/0.9545 & 30.43/0.9301 & 27.29/0.8803 \\
			\hline
		\end{tabular}
		\begin{tablenotes}
			\scriptsize
			\item NL: Non-Linearity; C: Context. $\Delta$ indicates increased parameters compared to the SegNet-DnCNN baseline. GFLOPs are measured on a $224\times224\times1$ input.
		\end{tablenotes}
	\end{threeparttable}
\end{table*}

\subsubsection{Network Architecture and Implementation Details}

\textit{Network Architecture}. We use the 17-layer DnCNN model~\cite{zhang2017beyond}, implemented by PyTorch.  To enable the use of IndexNet, we modify DnCNN to a SegNet-like architecture with $3$ downsampling and upsampling stages (the input image size is $40\times40$). The number of layers remains the same to ensure a relatively fair comparison. Fig.~\ref{fig:dncnn} illustrates the original  DnCNN  and our modified architecture. The first $9$ layers follow VGG-16 except
that
the first layer is a single-channel input, and the rest are $7$ decoding layers formed by unpooling and convolution and the final prediction layer. All convolutional operations
use
$3\times 3$ kernels. To incorporate IndexNet, it is straightforward to replace max pooling and unpooling with $\mathcal{IP}$ and $\mathcal{IU}$.

\textit{Training Details}. We follow the same experimental configurations used in~\cite{zhang2017beyond}. At each epoch, $40\times40$ image patches are cropped from multiple scales ($0.7, 0.8, 0.9, 1$) with a stride of $10$ and are added with Gaussian noise of a certain noise level ($\sigma=15,25,$ or $50$); image patches are further augmented with random flipping and random rotation. This results in around $240,000$ training samples. $\ell_2$ loss is used. All networks are trained from scratch with a batch size of $128$. Model parameters are initialized with the improved Xavier~\cite{he2015delving}. The Adam optimizer is also used. Parameters are updated with $60$ epochs. The learning rate is initially set to $0.001$ and reduced by $10\times$ at the $45$-th and $55$-th epoch, respectively.

\subsubsection{Results on the BSD68 and Set12 Datasets}

Apart from the DnCNN baseline, we also report the performance of our modified DnCNN-SegNet with max pooling and unpooling. Furthermore, to compare IndexNet against CARAFE, we build three additional baselines where CARAFE is combined with different downsampling strategies, including max pooling, average pooling, and stride-2  convolutions, denoted by DnCNN-max-carafe, DnCNN-avg-carafe, and DnCNN-conv-carafe, respectively. Results are shown in Table~\ref{tab:indexnet_denoising}. It can be observed that, simply downsampling with max pooling and upsampling by unpooling as in DnCNN-SegNet lead to significant drops in both PSNR (generally $>1 dB$) and SSIM ($>0.1$). This suggests
that
spatial information plays an important role in image denoising.
Denoising
is content-irrelevant (the model is unaware of regions coming from the foreground or the background).
Downsampling without recording sufficient spatial information (only the boundary information is not sufficient) impedes the model from recovering the appearance and the structure in the original image. This is particularly true for baselines adopting CARAFE. Since CARAFE %
applies
blind upsampling, no spatial information is transferred during upsampling, which may lead to inferior results.
Interestingly, after IndexNet is inserted into downsampled DnCNN, the loss of PSNR and SSIM is effectively compensated. The compensation behaviors can be observed from almost all types of IndexNet, except the two cases in Modelwise O2O DINs with nonlinearity. The poor performance of Modelwise O2O DINs may attribute to the insufficient modeling ability, particularly when $\sigma=50$.
Nonlinearity and weak context generally have a positive effect on image denoising, and the effectiveness of different IndexNets is similar. Hence, Shared Stagewise O2O DINs appear to be a preferred choice due to slightly increased parameters and negligible extra computation costs.

\subsection{Semantic Segmentation}

Here we further evaluate IndexNet on
semantic segmentation. Semantic segmentation
aims to predict a dense labeling map for each image where each pixel is labeled into one category.
Since the %
FCNs were introduced
\cite{long2015fully}, FCN-based encoder-decoder architectures have been studied extensively~\cite{chen2017deeplab,
lin2017refine,badrinarayanan2017segnet,chen18v3}.
Efforts have been spent on how to encode contextual information,
We
use SegNet~\cite{badrinarayanan2017segnet}
 as our baseline because IndexNet is primarily inspired by the unpooling operator in SegNet.
 We follow the experimental setting in~\cite{badrinarayanan2017segnet} and report performance on the SUN RGB-D~\cite{song2015sun} dataset.
 We use RGB as the input (depth is not used). The standard mean Intersection-over-Union (mIoU) is used as the evaluation metric.
 In addition,
 we
also compare against the recent
 UperNet~\cite{xiao2018unified}. We evaluate UperNet on the ADE20K dataset~\cite{zhou2017scene}.

\begin{figure}[!tb]
	\setlength{\abovecaptionskip}{10pt}
	\centering
	\includegraphics[width=\linewidth,angle=0]{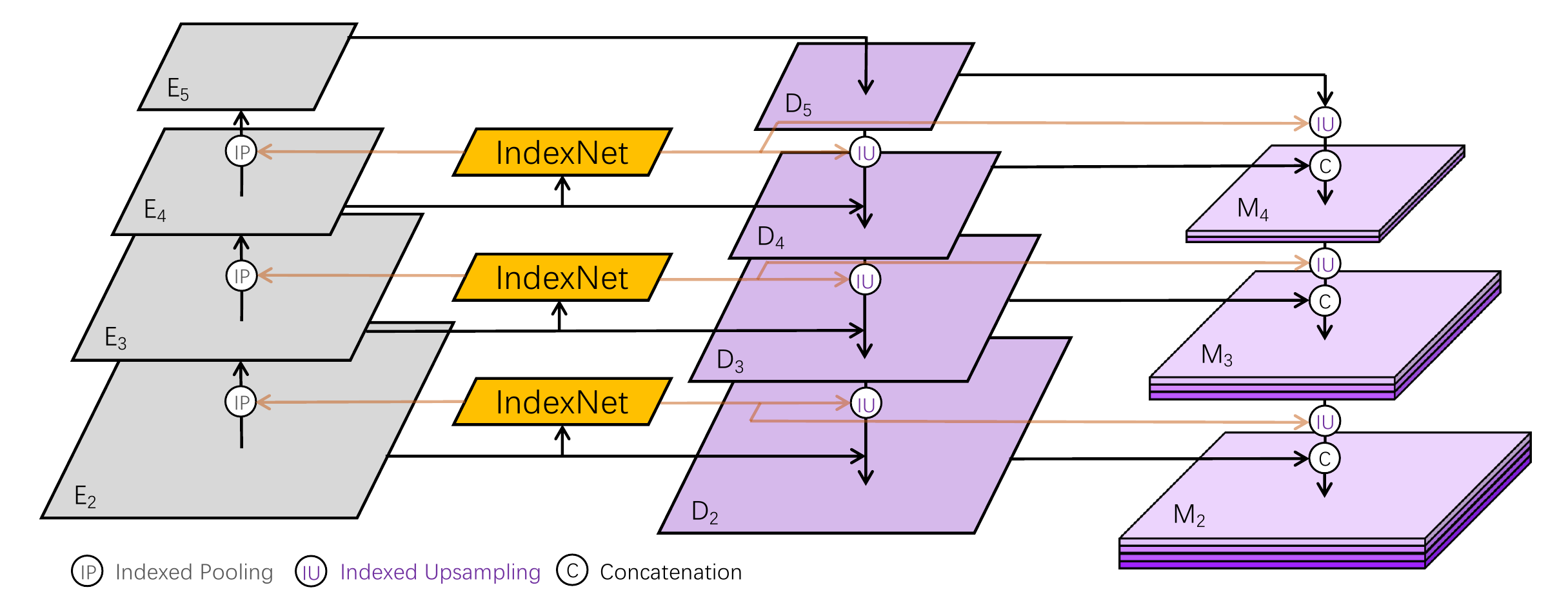}\vspace{-10pt}
	\caption{IndexNet-guided feature pyramid network and multi-level feature fusion.}
	\label{fig:indexnet-fpn-mff}
\end{figure}

\subsubsection{Network Architecture and Implementation Details}

\textit{Network Architecture.} The architecture of SegNet
employs the first $13$ layers of the VGG-16 model pretrained on ImageNet as the encoder. The decoder
uses
unpooling for upsampling. Each unpooling layer is followed by the same number of convolutional layers as in the corresponding encoder stage. Overall, SegNet has $5$ downsampling and $5$ upsampling stages. Convolutional layers in the decoding stage mainly play a role to smooth the feature maps generated by unpooling. To insert IndexNet, the only modification is to replace max pooling and unpooling layers with $\mathcal{IP}$ and $\mathcal{IU}$, respectively, which is straightforward.

UperNet builds upon the idea of Pyramid Pooling Module (PPM)~\cite{zhao2017pyramid} and Feature Pyramid Network (FPN)~\cite{lin2017feature}. UperNet also implements a Multi-level Feature Fusion (MFF) module that fuses multi-resolution feature maps by concatenation. In FPN, downsampling is implemented by 2-stride convolution, and upsampling uses bilinear interpolation. It produces four feature levels $\{D_2, D_3, D_4, D_5\}$ with output strides of $\{4,8,16,32\}$, conditioned on the encoder features $\{E_2, E_3, E_4, E_5\}$. MFF further fuses four levels of features and generates the output $M_2$ with an output stride of $4$. To insert IndexNet, three IndexNet blocks can be inserted into the encoder to generate index maps to guide upsampling. The same index maps can also be used in MFF in a sequential upsampling manner to fuse features, as shown in Fig.~\ref{fig:indexnet-fpn-mff}. Note that, in theory IndexNet can also be applied to PPM, because PPM itself has internal downsampling and upsampling stages. However, we discourage the use of IndexNet in PPM, because it will significantly increase parameters (due to mixed downsampling/upsampling rates). In this case blind upsampling such as NN/bilinear interpolation may be a better choice.

\textit{Training Details.} On the SUN RGB-D dataset, the VGG-16 model pretrained on ImageNet with BN layers is used. We employ the standard data augmentation strategies: random scaling, random cropping $320\times320$ sub-images, and random horizontal flipping. We learn the model with the standard softmax loss. Encoder parameters are pretrained on ImageNet. All other parameters are initialized with the improved Xavier~\cite{he2015delving}. The SGD optimizer~\cite{kingma2015adam} is used with a momentum of $0.9$ and a weight decay of $0.0001$. We train the model with a batch size of $16$ for $300$ epochs (around $90,000$ iterations). The learning rate is initially set to $0.01$ and reduced by $10\times$ at the $250$-th and $280$-th epoch, respectively. The BN layers of the encoder are fixed.

On the ADE20K benchmark, we use the MobileNetV2 pretrained on ImageNet as the encoder and UperNet the decoder. Due to limited computational resources, only this setting enables us to train a model on $4$ GPUs with a batch size of $16$ following the official implementation of UperNet and provided experimental settings.\footnote{\url{https://github.com/CSAILVision/semantic-segmentation-pytorch}}

\begin{figure*}[!tb]
	\captionsetup{font=small,singlelinecheck=true}
	\centering
	\includegraphics[width=.8\textwidth,angle=0]{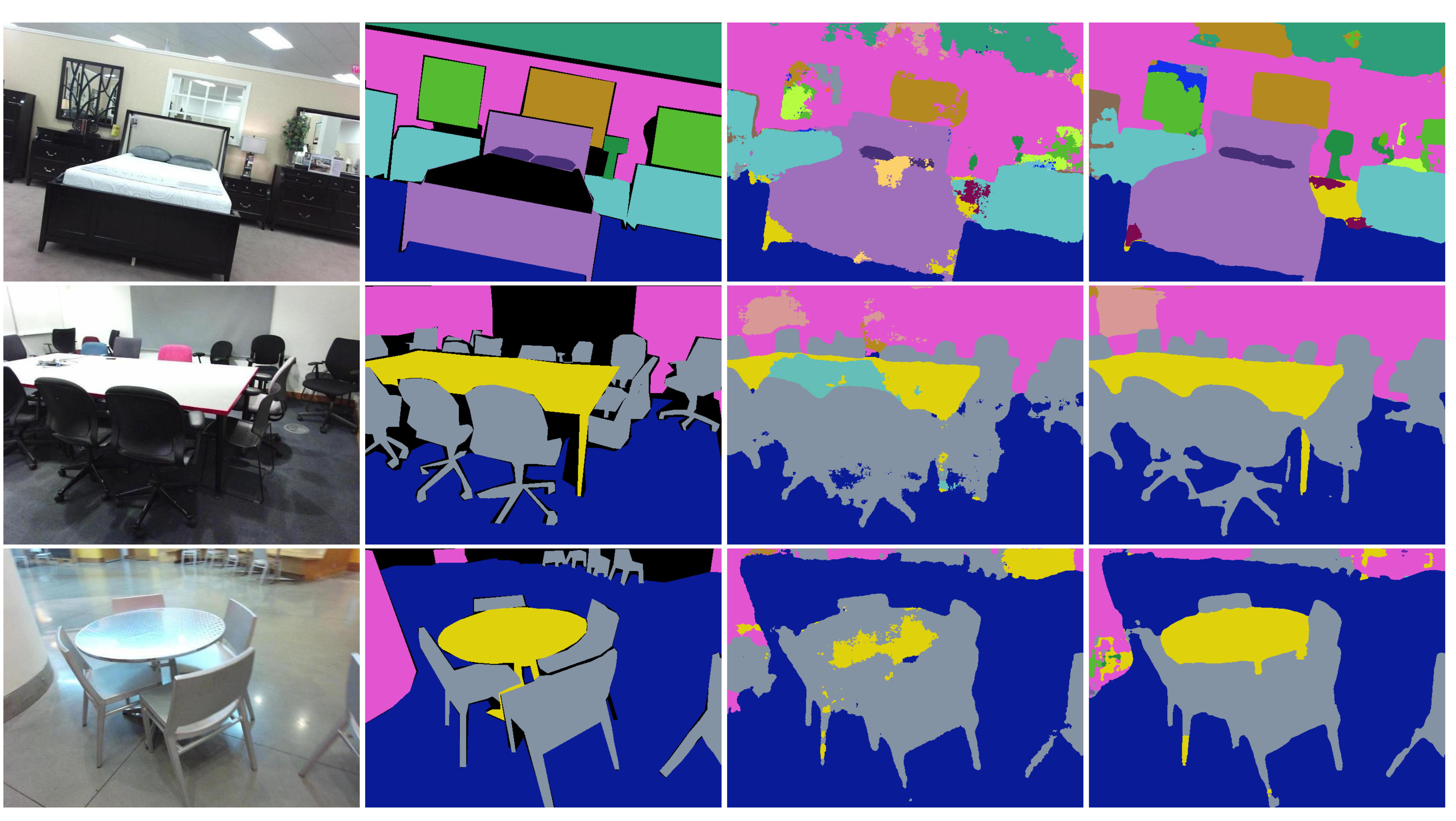}\vspace{-10pt}
	\caption{Semantic segmentation  results on the SUNRGB-D dataset. From left to right, the original image, ground-truth, SegNet, and ours (Shared Stagewise O2O DIN with `Nonlinearity').}
	\label{fig:segmentation_results}
\end{figure*}

\subsubsection{Results on the SUN RGB-D Dataset}

We report the
results
in Table~\ref{tab:indexnet_segmentation}. All index networks
show
improvements over the baseline, among which Modelwise and Shared Stagewise O2O DINs improve the baseline with few extra parameters and GFLOPs. Compared with other types of IndexNet, M2O DINs and HINs (particularly under the setting of ``Nonlinearity+Context'') increase many parameters and GFLOPs but do not exhibit
clear advantages.

From the qualitative results shown in Fig.~\ref{fig:segmentation_results}, we can see that the improvement comes from the ability of IndexNet suppressing fractured predictions that frequently appears in the baseline SegNet.
IndexNet seemingly does better in producing predictions at boundaries.

Notice that, in contrast to the behaviour in matting and denoising, CARAFE significantly enhances the performance in segmentation ($32.47\rightarrow36.30$), outperforming
IndexNet. We observe that CARAFE tends to produce consistent region-wise predictions. A plausible explanation is that, CARAFE is designed in a way to tackle region-sensitive tasks such as semantic segmentation where region-wise matching between predictions and ground truths matters, while IndexNet prefers detail-sensitive tasks like image matting where errors come from detail-rich
regions.

\begin{table}[!t]
	\centering
	\addtolength{\tabcolsep}{5pt}
	\renewcommand\arraystretch{1.0}
	\caption{Performance on the SUN RGB-D Dataset.}
	\label{tab:indexnet_segmentation}
	\vspace{-5pt}
	\begin{threeparttable}
		\begin{tabular}{c|c|c|c|c}
			\hline
			\multicolumn{2}{c|}{Method} & \#Param. & GFLOPs & mIoU\\
			\hline
			\multicolumn{2}{c|}{SegNet~\cite{badrinarayanan2017segnet}} & 24.96M & 24.76 & 32.47\\
			\multicolumn{2}{c|}{SegNet-carafe} & 25.35M & 25.75 & \textbf{36.30}\\
			\hline
			NL & C & $\Delta$ \\
			\hline
			& & \multicolumn{3}{c}{HINs}\\
			\hline
			& 							 				& +23.55K & 24.79 & 33.25\\
			\checkmark 	& 			 	 				& +4.90M  & 26.40 & 33.11\\
			\checkmark 	& \checkmark 	 				& +19.55M & 31.28 & 33.31\\
			\hline
			& & \multicolumn{3}{c}{Modelwise O2O DINs}\\
			\hline
			& 				 							& +16  & 24.76 & 33.18\\
			\checkmark 	& 								& +56  & 24.76 & 33.70\\
			\checkmark 	& \checkmark 					& +152 & 24.77 & 33.26\\
			\hline
			& & \multicolumn{3}{c}{Shared Stagewise O2O DINs}\\
			\hline
			& 								 			& +80  & 24.76 & 33.26\\
			\checkmark	& 								& +280 & 24.76 & 33.97\\
			\checkmark	& \checkmark 					& +760 & 24.77 & 33.41\\
			\hline
			& & \multicolumn{3}{c}{Unshared Stagewise O2O DINs}\\
			\hline
			&            					            & +0.02M & 24.79 & 33.27\\
			\checkmark 	&            		            & +0.08M & 24.82 & 33.59\\
			\checkmark 	& \checkmark 		            & +0.22M & 24.96 & 33.50\\
			\hline
			& & \multicolumn{3}{c}{M2O DINs}\\
			\hline
			&            				                & +9.76M  & 28.02 & 33.28\\
			\checkmark 	&            	                & +24.44M & 32.90 & 33.51\\
			\checkmark 	& \checkmark 		            & +83.02M & 52.42 & 33.48\\
			\hline
		\end{tabular}
		\begin{tablenotes}
			\scriptsize
			\item NL: Non-Linearity; C: Context. $\Delta$ indicates increased parameters compared to the SegNet baseline. GFLOPs are measured on a $224\times224\times3$ input.
		\end{tablenotes}
	\end{threeparttable}
\end{table}

\begin{table}[!t]
	\centering
	\addtolength{\tabcolsep}{0pt}
	\renewcommand\arraystretch{1.0}
	\caption{Performance on the ADE-20K Dataset}
	\label{tab:indexnet_segmentation_2}
	\vspace{-5pt}
	\begin{threeparttable}
		\begin{tabular}{c|c|c|c}
			\hline
			FPN & MFF & mIoU & Pixel Accuracy ($\%$)\\
			\hline
			Bilinear & Bilinear & 37.08 & 78.29\\
			IndexNet & Bilinear & 36.25 & 78.23\\
			Bilinear & IndexNet & 36.90 & 78.27\\
			IndexNet & IndexNet & 37.62 & 78.29\\
			CARAFE   & Bilinear & 37.76 & 78.81\\
			Bilinear & CARAFE   & 38.03 & 78.51\\
			CARAFE   & CARAFE   & \textbf{38.31} & \textbf{78.90}\\
			\hline
		\end{tabular}
		\begin{tablenotes}
			\scriptsize
			\item
			MFF: Multi-level Feature Fusion. Only HINs (`Linear') are evaluated due to varied decoder feature dimensionality. The best performance is boldfaced.
		\end{tablenotes}
	\end{threeparttable}
\end{table}

\subsubsection{Results on the ADE20K Dataset}
Here we conduct
ablative studies to highlight the role of upsampling in UperNet. Both IndexNet and CARAFE are considered. In addition to the full replacement of upsampling operators following Fig.~\ref{fig:indexnet-fpn-mff}, we also replace bilinear upsampling either in FPN or in MFF with IndexNet/CARAFE. Results are shown in Table~\ref{tab:indexnet_segmentation_2}. It can be observed that, IndexNet
improves
UperNet ($37.08\rightarrow37.62$) only when bilinear upsampling in FPN and MFF is simultaneously replaced. However, when only one component is modified, IndexNet even leads to negative results. This suggests that the guided information should be used consistently in the decoder. In addition, CARAFE also works better than IndexNet, showing that
spatial information may not play a critical
role
in semantic segmentation.

\subsection{Monocular Depth Estimation}

Estimating per-pixel depth from a single image
is challenging
because one needs to recover 3D information from a 2D plane.
With deep learning, significant progress has been witnessed %
\cite{%
liu2015learning,
xian2018monocular,wofk2019fastdepth}.
We use the recent
FastDepth~\cite{wofk2019fastdepth} as our baseline.
We compare the performance with/without IndexNet on the NYUv2 dataset~\cite{silberman2012indoor} with the official train/test split. To be in consistent with~\cite{wofk2019fastdepth}, the following metrics are used to quantify the performance:
\begin{itemize}
	\item root mean square error (rms): $\sqrt { \frac { 1 }{ T } \sum _{ i=1 }^{ T }{ { \left( { d }_{ i }-{ g }_{ i } \right)  }^{ 2 } }  } $;
	\item accuracy with threshold $th$: percentage (\%) of $d_1$, $\rm s.t.$
	$\max { \left( \frac { { d }_{ 1 } }{ { g }_{ 1 } } ,\frac { { g }_{ 1 } }{ { d }_{  } }  \right)  } =\delta_{1} <$ $th$.
\end{itemize}

\subsubsection{Network Architecture and Implementation Details}

\begin{figure}[!tb]
	\captionsetup{font=small,singlelinecheck=true}
	\setlength{\abovecaptionskip}{10pt}
	\centering
	\includegraphics[width=3.5in,angle=0]{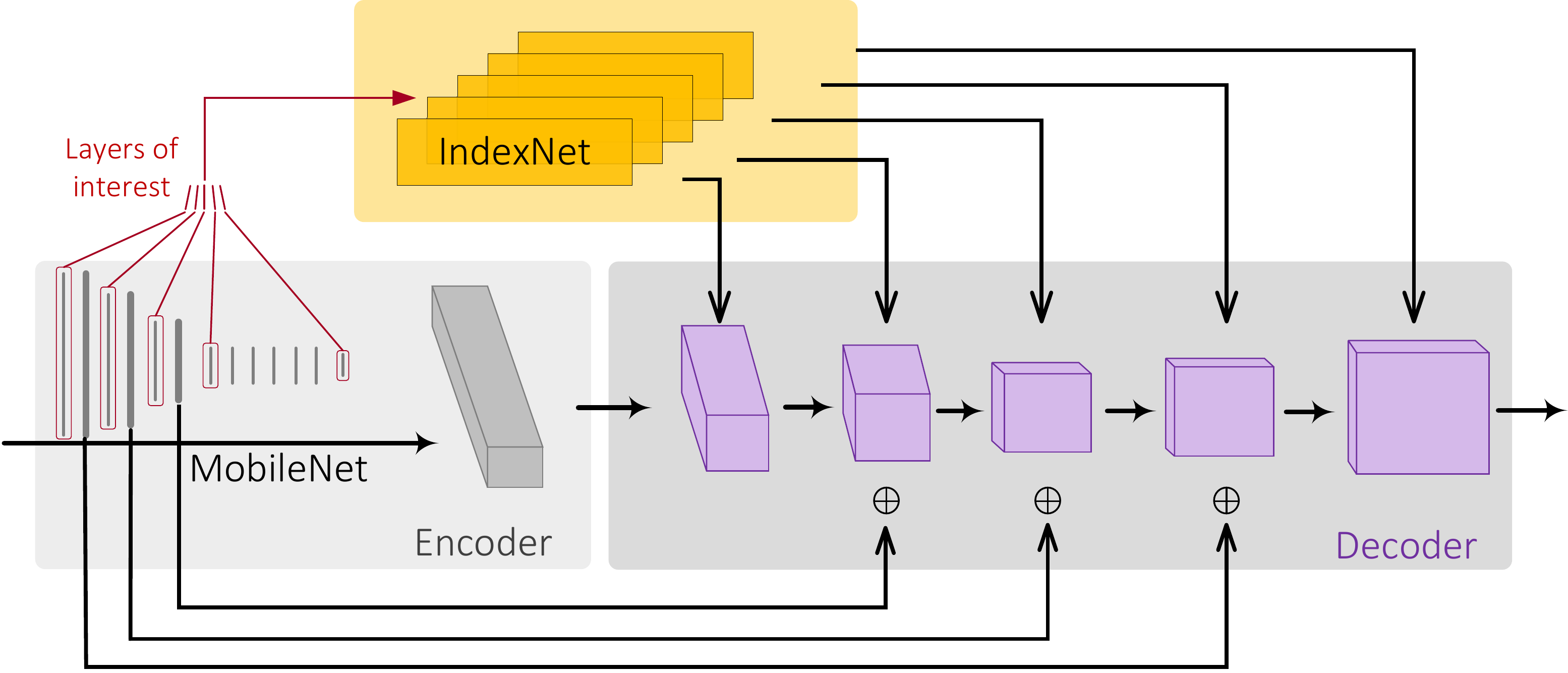}\vspace{-8pt}
	\caption{Our modified FastDepth~\cite{wofk2019fastdepth} architecture.}
	\label{fig:fastdepth_architecture}
\end{figure}

\textit{Network Architecture.} FastDepth is an
encoder-decoder architecture, with MobileNet as its backbone.
Here we choose the best upsampling option suggested by the authors~\cite{wofk2019fastdepth} where upsampling is implemented by $\times2$ NN interpolation and $5\times5$ convolution. Hence, our baseline is FastDepth-NNConv5: downsampling with 2-stride convolution and upsampling via NN interpolation. We also modify this baseline by changing the stride-2 convolution to be stride-1  followed by max-pooling, named as FastDepth-P-NNConv5.
Fig.~\ref{fig:fastdepth_architecture} shows how we insert IndexNet into FastDepth. Similar to the modifications applied to the matting network, stride-2  convolution layers in the encoder are changed to be stride-1, followed by $\mathcal{IP}$, and the NN interpolation in the decoder is replaced with $\mathcal{IU}$. To compare IndexNet with CARAFE, we build two additional baselines: FastDepth-carafe and FastDepth-P-carafe, where NNConv5 is modified to CARAFE in FastDepth-NNConv5 and FastDepth-P-NNConv5.

\textit{Training Details.} We follow similar training settings used by FastDepth~\cite{wofk2019fastdepth}. $\ell_1$ loss is used. Random rotation, random scaling and random horizontal flipping are used for data augmentation. The initial learning rate is set to $0.01$ and reduced by $\times10$ every $5$ epochs. The SGD optimizer is used with a momentum of $0.9$ and a weight decay of $0.0001$. Encoder weights are pretrained on ImageNet~\cite{deng2009imagenet}. A batch size of $16$ is used to train the network for $30$ epochs in total.

\begin{table}[!t]
	\captionsetup{font=small,singlelinecheck=true}
	\centering
	\addtolength{\tabcolsep}{-1pt}
	\renewcommand\arraystretch{1.0}
	\caption{Performance of FastDepth~\cite{wofk2019fastdepth} on the NYUDv2 Dataset.}
	\label{tab:indexnet_fastdepth}
	\vspace{-5pt}
	\begin{threeparttable}
		\begin{tabular}{>{\centering}p{1.1cm}|>{\centering}p{1.45cm}|c|c|c|c}
			\hline
			\multicolumn{2}{c|}{Method} & \#Param. & GFLOPs & rms & $\delta_{1}<1.25$\\
			\hline

			\hline
			\multicolumn{2}{c|}{FastDepth-NNConv5} & 3.96M & 0.69 & 0.567 & 0.781\\
			\multicolumn{2}{c|}{FastDepth-P-NNConv5} & 3.96M & 1.01 & 0.577 & 0.778 \\
			\multicolumn{2}{c|}{FastDepth-carafe} & 4.31M & 1.63 & 0.558 & \textbf{0.790} \\
			\multicolumn{2}{c|}{FastDepth-P-carafe} & 4.31M & 1.96 & 0.571 & 0.782 \\
			\hline
			NL & C & $\Delta$\\
			\hline
			& & \multicolumn{4}{c}{HINs}\\
			\hline
			& & +31.23K & 1.03 & 0.566 & 0.784\\
			\checkmark & & +11.17M & 2.65 & 0.565 & 0.786\\
			\checkmark & \checkmark & +44.62M & 7.53 & 0.559 & 0.787\\
			\hline
			& & \multicolumn{4}{c}{Modelwise O2O DINs}\\
			\hline
			& & +16 & 1.02 & 0.569 & 0.778\\
			\checkmark & & +56 & 1.02 & 0.568 & 0.785 \\
			\checkmark & \checkmark & +152 & 1.02 & 0.564 & 0.786\\
			\hline
			& & \multicolumn{4}{c}{Shared Stagewise O2O DINs}\\
			\hline
			& & +80 & 1.02 & 0.562 & 0.783\\
			\checkmark & & +280 & 1.02 & 0.565 & 0.786\\
			\checkmark & \checkmark & +760 & 1.02 & 0.567 & 0.783\\
			\hline
			& & \multicolumn{4}{c}{Unshared Stagewise O2O DINs}\\
			\hline
			& & +31.23K & 1.03 & \textbf{0.556} & 0.789\\
			\checkmark & & +0.11M & 1.06 & 0.564 & 0.786\\
			\checkmark & \checkmark & +0.30M & 1.16 & 0.562 & 0.788\\
			\hline
			& & \multicolumn{4}{c}{M2O DINs}\\
			\hline
			& & +22.30M & 4.27 & 0.563 & 0.783\\
			\checkmark & & +55.78M & 9.15 & 0.562 & 0.786\\
			\checkmark & \checkmark & +189.57M & 28.67 & 0.565 & 0.787\\
			\hline
		\end{tabular}
		\begin{tablenotes}
			\scriptsize
			\item NL: Non-Linearity; C: Context. $\Delta$ indicates increased parameters compared to the standard FastDepth baseline. GFLOPs are measured on a $224\times224\times3$ input. The best performance is boldfaced.
		\end{tablenotes}
	\end{threeparttable}
	\vspace{-10pt}
\end{table}

\begin{table*}[!t]
    \centering
	\addtolength{\tabcolsep}{6pt}
	\renewcommand\arraystretch{1.0}
	\caption{Performance of Hu \textit{et al.}~\cite{hu2019revisiting} on the NYUDv2 dataset}
    \label{tab:indexnet_depth_2}
	\vspace{-5pt}
	\begin{threeparttable}
    \begin{tabular}{c|c c c|c c c}
    \hline
      & rms & rel & log10 & $\delta<1.25$ & $\delta<1.25^2$ & $\delta<1.25^3$ \\
      \hline
       Hu \MakeLowercase{\textit{et al.}}~\cite{hu2019revisiting}  &  0.558 & 0.129 & 0.055 & 0.837 & \textbf{0.968} & \textbf{0.992} \\
        Hu \MakeLowercase{\textit{et al.}} + IndexNet &  \textbf{0.554} & \textbf{0.128} & \textbf{0.054} & \textbf{0.843} & \textbf{0.968} & \textbf{0.992} \\
        \hline
    \end{tabular}
    \begin{tablenotes}
			\scriptsize
			\item Only HIN (`Nonlinear+Context') is evaluated due to varied feature dimensionality of decoder and multi-level feature fusion. \textit{rel} and \textit{log10} denote the average relative error and average $\log_{10}$ error~\cite{xian2018monocular}, respectively. The best performance is boldfaced.
		\end{tablenotes}
    \end{threeparttable}
\end{table*}

\subsubsection{Results on the NYUDv2 Dataset}

\begin{figure*}[!tb]
	\captionsetup{font=small,singlelinecheck=true}
	\setlength{\abovecaptionskip}{10pt}
	\centering
	\includegraphics[width=0.65\textwidth,angle=0]{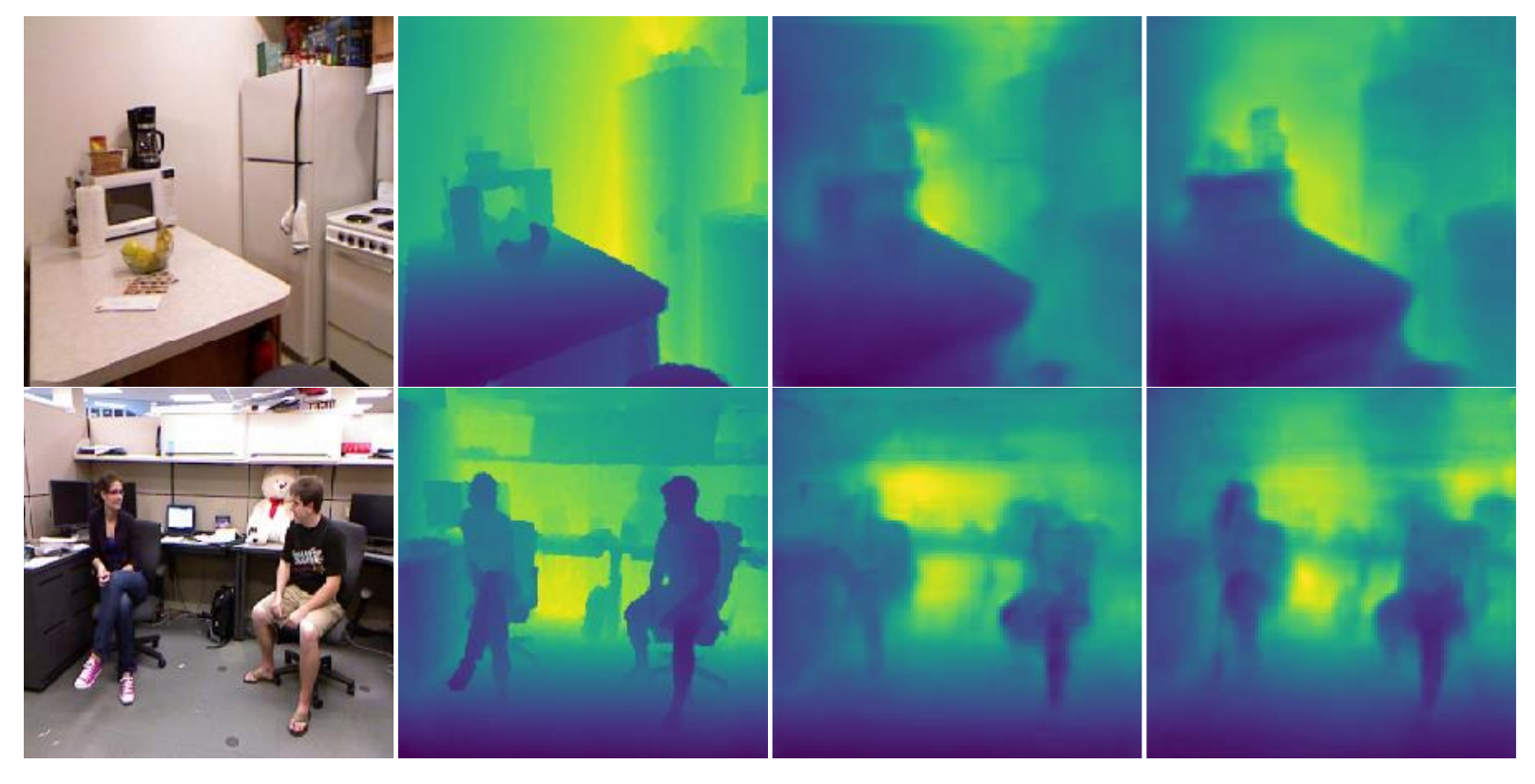}\vspace{-8pt}
	\caption{Qualitative results on the NYUDv2 dataset. From left to right, the original image, ground-truth, FastDepth-NNConv5, and ours (Unshared Stagewise O2O DIN with `Linear').}
	\label{fig:fastdepth}
\end{figure*}

We report the results  in Table~\ref{tab:indexnet_fastdepth}. We observe that
almost
all types of IndexNet improve the performance compared to the baselines except for the most light-weight design---linear Modelwise O2O DIN.
It may be because
only $16$ parameters are not sufficient to model local variations of high-dimensional feature maps.
Note that, Unshared Stagewise O2O DINs (with only linear mappings)
shows
clear improvements with only
slightly
increased parameters.
HINs and M2O DINs increase a large amount of parameters and floating-point calculations because of the high dimensionality of feature maps, while the improved performance is not proportional to such a high cost.
Some qualitative results are further illustrated in Fig.~\ref{fig:fastdepth}. We observe that IndexNet exhibits better boundary delineation than the baseline, e.g., the edge of the desk, and the contour of the woman. Moreover, CARAFE achieves comparable performance against IndexNet in this task.

In addition, we report the results of applying IndexNet to a state-of-the-art model~\cite{hu2019revisiting} in Table~\ref{tab:indexnet_depth_2}.
We modify the usage of IndexNet here by taking the same feature fusion strategies shown in Fig.~\ref{fig:indexnet-fpn-mff}. Other implementation details and evaluations are kept consistent with~\cite{hao2019indexnet}. Compared with the baseline, IndexNet
shows
improvement in the first four metrics.

\subsection{Insights Towards Good Practices}
\label{ssec:practices}

As a summary of our evaluations, here we provide some guidelines %
for
using  guided/blind upsampling:
\begin{enumerate}
    \item

    In detail-sensitive tasks, such as image matting, image restoration, and edge detection, the spatial information is important.
    Thus guided upsampling may be preferred.

    \item
    Blind upsampling %
    may be used
    in the situation when computational %
    budget is
    limited because most blind upsampling operators are non-parametric and computationally efficient.

	\item

	In image matting, the best IndexNet configuration is ``M2O DINs+Nonlinearity+Context''. This configuration is also true for the
	image reconstruction experiment
	and image denoising, where M2O DINs exhibit the best performance and the most stable behavior, respectively. Hence, the capacity of IndexNet is closely related to the complexity of local patterns. M2O DINs is %
	preferred
	in a detail- or boundary-sensitive task, but one should also be aware of the increased model parameters and computation costs, especially when the feature maps are high-dimensional.
	\item If one prefers a flexible decoder design, e.g., squeezing/enlarging the dimensionality of the decoder feature map, HINs are good choices, because DINs only generate index maps whose dimensionality is identical to the input feature map.
	\item
	For real-time applications,
	Shared Stagewise O2O DINs are the first choices. Model parameters increased by Shared Stagewise O2O DINs are comparable to Modelwise O2O DINs, and the extra GFLOPs are also negelectable. Shared Stagewise O2O DINs, however, always work better than Modelwise O2O DINs
	 for applications considered in this work. It  implies that each upsampling stage should learn a stage-specific index function;
	\item It is worth noting that, the current implementation of IndexNet has some limitations. Currently IndexNet only implements single-point upsampling---each upsampled feature point is only associated with a single point. In this sense, we may not simulate the behavior of bilinear interpolation where each upsampled point is affected by multiple points of a local region.
\end{enumerate}

\section{Conclusion}

Inspired by an observation in image matting, we
examine
the role of indices and present a unified
view
of upsampling operators using the notion of index functions. We show that an index function can be learned within a proposed index-guided encoder-decoder framework. In this framework, indices are learned with a flexible network module termed IndexNet, and are used to guide downsampling and upsampling using $\mathcal{IP}$ and $\mathcal{IU}$. IndexNet itself is also a sub-framework that can be designed depending on the task at hand. We %
investigate five index networks,
and  demonstrate their effectiveness on four
dense prediction tasks.
We believe that IndexNet is an important step towards generic upsampling operators for deep networks.

\ifCLASSOPTIONcaptionsoff
  \newpage
\fi

\begin{IEEEbiography}[{\includegraphics[width=1in,height=1.25in,clip,keepaspectratio]{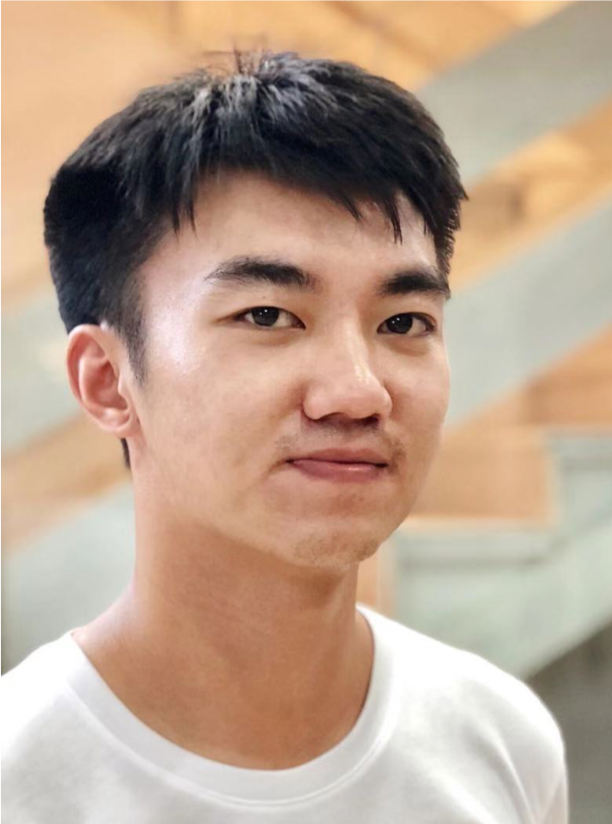}}]{Hao Lu}
	received the Ph.D. degree from Huazhong University of Science and Technology, Wuhan, China, in 2018.
	He is currently a Postdoctoral Fellow at School of Computer Science, The University of Adelaide, Australia. His research interests include computer vision and machine learning. He is currently working on dense prediction problems.
\end{IEEEbiography}
\vskip 0pt plus -1fil

\begin{IEEEbiography}[{\includegraphics[width=1in,height=1.25in,clip,keepaspectratio]{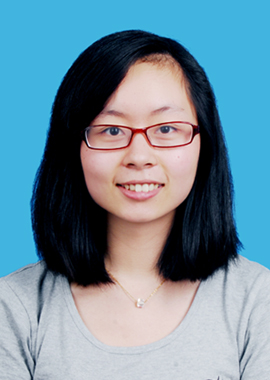}}]{Yutong Dai}
    received
    the M.Sc. degree from Southeast University, Nanjing, in 2018.
    She
    is
    currently
    pursuing the Ph.D.\  degree
    at The University of Adelaide, Australia. Her research interests include computer vision and deep learning. She is currently working on image matting.
\end{IEEEbiography}
\vskip 0pt plus -1fil

\begin{IEEEbiographynophoto}
    {Chunhua Shen}
    is a Professor of Computer Science at The University of Adelaide, Australia.
\end{IEEEbiographynophoto}

    \vskip 0pt plus -1fil

\begin{IEEEbiographynophoto}
{Songcen Xu}
    received the Ph.D. degree in electronic engineering from the University of York, U.K. in 2015.
    He is
    with Noah's Ark Lab, Huawei Technologies.
\end{IEEEbiographynophoto}
\vskip 0pt plus -1fil


\begin{thebibliography}{10}
\providecommand{\url}[1]{#1}
\csname url@samestyle\endcsname
\providecommand{\newblock}{\relax}
\providecommand{\bibinfo}[2]{#2}
\providecommand{\BIBentrySTDinterwordspacing}{\spaceskip=0pt\relax}
\providecommand{\BIBentryALTinterwordstretchfactor}{4}
\providecommand{\BIBentryALTinterwordspacing}{\spaceskip=\fontdimen2\font plus
\BIBentryALTinterwordstretchfactor\fontdimen3\font minus
  \fontdimen4\font\relax}
\providecommand{\BIBforeignlanguage}[2]{{%
\expandafter\ifx\csname l@#1\endcsname\relax
\typeout{** WARNING: IEEEtran.bst: No hyphenation pattern has been}%
\typeout{** loaded for the language `#1'. Using the pattern for}%
\typeout{** the default language instead.}%
\else
\language=\csname l@#1\endcsname
\fi
#2}}
\providecommand{\BIBdecl}{\relax}
\BIBdecl

\bibitem{zeiler2014visualizing}
M.~D. Zeiler and R.~Fergus, ``Visualizing and understanding convolutional
  networks,'' in \emph{Proc. European Conference on Computer Vision (ECCV)},
  2014, pp. 818--833.

\bibitem{long2015fully}
J.~Long, E.~Shelhamer, and T.~Darrell, ``Fully convolutional networks for
  semantic segmentation,'' in \emph{Proc. IEEE Conference on Computer Vision
  and Pattern Recognition (CVPR)}, 2015, pp. 3431--3440.

\bibitem{badrinarayanan2017segnet}
V.~Badrinarayanan, A.~Kendall, and R.~Cipolla, ``{SegNet}: {A} deep
  convolutional encoder-decoder architecture for image segmentation,''
  \emph{IEEE Transactions on Pattern Analysis and Machine Intelligence},
  vol.~39, no.~12, pp. 2481--2495, 2017.

\bibitem{shi2016real}
W.~Shi, J.~Caballero, F.~Husz{\'a}r, J.~Totz, A.~P. Aitken, R.~Bishop,
  D.~Rueckert, and Z.~Wang, ``Real-time single image and video super-resolution
  using an efficient sub-pixel convolutional neural network,'' in \emph{Proc.
  IEEE Conference on Computer Vision and Pattern Recognition (CVPR)}, 2016, pp.
  1874--1883.

\bibitem{lin2017refine}
G.~Lin, A.~Milan, C.~Shen, and I.~Reid, ``{RefineNet}: {Multi-path} refinement
  networks for high-resolution semantic segmentation,'' in \emph{Proc. IEEE
  Conference on Computer Vision and Pattern Recognition (CVPR)}, 2017, pp.
  1925--1934.

\bibitem{chen18v3}
L.-C. Chen, Y.~Zhu, G.~Papandreou, F.~Schroff, and H.~Adam, ``Encoder-decoder
  with atrous separable convolution for semantic image segmentation,'' in
  \emph{Proc. European Conference on Computer Vision (ECCV)}, 2018.

\bibitem{xu2017deep}
N.~Xu, B.~Price, S.~Cohen, and T.~Huang, ``Deep image matting,'' in \emph{Proc.
  IEEE Conference on Computer Vision and Pattern Recognition (CVPR)}, 2017, pp.
  2970--2979.

\bibitem{kraska2018case}
T.~Kraska, A.~Beutel, E.~H. Chi, J.~Dean, and N.~Polyzotis, ``The case for
  learned index structures,'' in \emph{Proc. International Conference on
  Management of Data}, 2018, pp. 489--504.

\bibitem{sandler2018mobilenetv2}
M.~Sandler, A.~Howard, M.~Zhu, A.~Zhmoginov, and L.-C. Chen, ``Mobilenetv2:
  Inverted residuals and linear bottlenecks,'' in \emph{Proc. IEEE Conference
  on Computer Vision and Pattern Recognition (CVPR)}, 2018, pp. 4510--4520.

\bibitem{zhang2017beyond}
K.~Zhang, W.~Zuo, Y.~Chen, D.~Meng, and L.~Zhang, ``Beyond a {G}aussian
  denoiser: residual learning of deep {CNN} for image denoising,'' \emph{IEEE
  Transactions on Image Processing}, vol.~26, no.~7, pp. 3142--3155, 2017.

\bibitem{roth2009fields}
S.~Roth and M.~J. Black, ``Fields of experts,'' \emph{International Journal of
  Computer Vision}, vol.~82, no.~2, p. 205, 2009.

\bibitem{song2015sun}
S.~Song, S.~P. Lichtenberg, and J.~Xiao, ``{SUN RGB-D}: {A RGB-D} scene
  understanding benchmark suite,'' in \emph{Proc. IEEE Conference on Computer
  Vision and Pattern Recognition (CVPR)}, 2015, pp. 567--576.

\bibitem{silberman2012indoor}
N.~Silberman, D.~Hoiem, P.~Kohli, and R.~Fergus, ``Indoor segmentation and
  support inference from rgbd images,'' in \emph{Proc. European Conference on
  Computer Vision (ECCV)}, 2012, pp. 746--760.

\bibitem{hao2019indexnet}
H.~Lu, Y.~Dai, C.~Shen, and S.~Xu, ``Indices matter: Learning to index for deep
  image matting,'' in \emph{Proc. IEEE/CVF International Conference on Computer
  Vision (ICCV)}, 2019, pp. 3266--3275.

\bibitem{odena2016deconvolution}
A.~Odena, V.~Dumoulin, and C.~Olah, ``Deconvolution and checkerboard
  artifacts,'' \emph{Distill}, vol.~1, no.~10, p.~e3, 2016.

\bibitem{osendorfer2014image}
C.~Osendorfer, H.~Soyer, and P.~Van Der~Smagt, ``Image super-resolution with
  fast approximate convolutional sparse coding,'' in \emph{Proc. International
  Conference on Neural Information Processing (ICONIP)}, 2014, pp. 250--257.

\bibitem{yang2019deeperlab}
T.-J. Yang, M.~D. Collins, Y.~Zhu, J.-J. Hwang, T.~Liu, X.~Zhang, V.~Sze,
  G.~Papandreou, and L.-C. Chen, ``{DeeperLab}: Single-shot image parser,''
  \emph{arXiv}, 2019.

\bibitem{jiaqi2019carafe}
J.~Wang, K.~Chen, R.~Xu, Z.~Liu, C.~C. Loy, and D.~Lin, ``{CARAFE}:
  Context-aware reassembly of features,'' in \emph{Proc. IEEE/CVF International
  Conference on Computer Vision (ICCV)}, 2019.

\bibitem{jaderberg2015spatial}
M.~Jaderberg, K.~Simonyan, A.~Zisserman \emph{et~al.}, ``Spatial transformer
  networks,'' in \emph{Advances in Neural Information Processing Systems
  (NIPS)}, 2015, pp. 2017--2025.

\bibitem{jia2016dynamic}
X.~Jia, B.~De~Brabandere, T.~Tuytelaars, and L.~V. Gool, ``Dynamic filter
  networks,'' in \emph{Advances in Neural Information Processing Systems
  (NIPS)}, 2016, pp. 667--675.

\bibitem{dai2017deformable}
J.~Dai, H.~Qi, Y.~Xiong, Y.~Li, G.~Zhang, H.~Hu, and Y.~Wei, ``Deformable
  convolutional networks,'' in \emph{Proc. IEEE International Conference on
  Computer Vision (ICCV)}, 2017, pp. 764--773.

\bibitem{mnih2014recurrent}
V.~Mnih, N.~Heess, A.~Graves \emph{et~al.}, ``Recurrent models of visual
  attention,'' in \emph{Advances in Neural Information Processing Systems
  (NIPS)}, 2014, pp. 2204--2212.

\bibitem{wang2017residual}
F.~Wang, M.~Jiang, C.~Qian, S.~Yang, C.~Li, H.~Zhang, X.~Wang, and X.~Tang,
  ``Residual attention network for image classification,'' in \emph{Proc. IEEE
  Conference on Computer Vision and Pattern Recognition (CVPR)}, 2017, pp.
  3156--3164.

\bibitem{hu2018squeeze}
J.~Hu, L.~Shen, and G.~Sun, ``Squeeze-and-excitation networks,'' in \emph{Proc.
  IEEE Conference on Computer Vision and Pattern Recognition}, 2018, pp.
  7132--7141.

\bibitem{woo2018cbam}
S.~Woo, J.~Park, J.-Y. Lee, and I.~So~Kweon, ``{CBAM}: Convolutional block
  attention module,'' in \emph{Proc. European Conference on Computer Vision
  (ECCV)}, 2018, pp. 3--19.

\bibitem{xiao2017fashion}
H.~Xiao, K.~Rasul, and R.~Vollgraf, ``Fashion-mnist: a novel image dataset for
  benchmarking machine learning algorithms,'' \emph{arXiv}, 2017.

\bibitem{chen2013knn}
Q.~Chen, D.~Li, and C.-K. Tang, ``{KNN} matting,'' \emph{IEEE Transactions on
  Pattern Analysis and Machine Intelligence}, vol.~35, no.~9, pp. 2175--2188,
  2013.

\bibitem{chuang2001bayesian}
Y.-Y. Chuang, B.~Curless, D.~H. Salesin, and R.~Szeliski, ``A bayesian approach
  to digital matting,'' in \emph{Proc. IEEE Conference on Computer Vision and
  Pattern Recognition (CVPR)}, 2001, pp. 264--271.

\bibitem{he2011global}
K.~He, C.~Rhemann, C.~Rother, X.~Tang, and J.~Sun, ``A global sampling method
  for alpha matting,'' in \emph{Proc. IEEE Conference on Computer Vision and
  Pattern Recognition (CVPR)}.\hskip 1em plus 0.5em minus 0.4em\relax IEEE,
  2011, pp. 2049--2056.

\bibitem{levin2008closed}
A.~Levin, D.~Lischinski, and Y.~Weiss, ``A closed-form solution to natural
  image matting,'' \emph{IEEE Transactions on Pattern Analysis and Machine
  Intelligence}, vol.~30, no.~2, pp. 228--242, 2008.

\bibitem{girshick2014rich}
R.~Girshick, J.~Donahue, T.~Darrell, and J.~Malik, ``Rich feature hierarchies
  for accurate object detection and semantic segmentation,'' in \emph{Proc.
  IEEE Conference on Computer Vision and Pattern Recognition (CVPR)}, 2014, pp.
  580--587.

\bibitem{krizhevsky2012imagenet}
A.~Krizhevsky, I.~Sutskever, and G.~E. Hinton, ``{ImageNet} classification with
  deep convolutional neural networks,'' in \emph{Advances in Neural Information
  Processing Systems (NIPS)}, 2012, pp. 1097--1105.

\bibitem{lin2014microsoft}
T.-Y. Lin, M.~Maire, S.~Belongie, J.~Hays, P.~Perona, D.~Ramanan,
  P.~Doll{\'a}r, and C.~L. Zitnick, ``Microsoft coco: Common objects in
  context,'' in \emph{Proc. European Conference on Computer Vision (ECCV)},
  2014, pp. 740--755.

\bibitem{everingham2010pascal}
M.~Everingham, L.~Van~Gool, C.~K. Williams, J.~Winn, and A.~Zisserman, ``The
  pascal visual object classes (voc) challenge,'' \emph{International Journal
  of Computer Vision}, vol.~88, no.~2, pp. 303--338, 2010.

\bibitem{rhemann2009perceptually}
C.~Rhemann, C.~Rother, J.~Wang, M.~Gelautz, P.~Kohli, and P.~Rott, ``A
  perceptually motivated online benchmark for image matting,'' in \emph{Proc.
  IEEE Conference on Computer Vision and Pattern Recognition (CVPR)}, 2009, pp.
  1826--1833.

\bibitem{deng2009imagenet}
J.~Deng, W.~Dong, R.~Socher, L.-J. Li, K.~Li, and L.~Fei-Fei, ``{ImageNet}: A
  large-scale hierarchical image database,'' in \emph{Proc. IEEE Conference on
  Computer Vision and Pattern Recognition (CVPR)}.\hskip 1em plus 0.5em minus
  0.4em\relax Ieee, 2009, pp. 248--255.

\bibitem{kingma2015adam}
D.~P. Kingma and J.~Ba, ``Adam: A method for stochastic optimization,'' in
  \emph{Proc. International Conference on Learning Representations (ICLR)},
  2015.

\bibitem{he2016deep}
K.~He, X.~Zhang, S.~Ren, and J.~Sun, ``Deep residual learning for image
  recognition,'' in \emph{Proc. IEEE Conference on Computer Vision and Pattern
  Recognition (CVPR)}, 2016, pp. 770--778.

\bibitem{ronneberger2015u}
O.~Ronneberger, P.~Fischer, and T.~Brox, ``{U-Net}: {C}onvolutional networks
  for biomedical image segmentation,'' in \emph{Proc. International Conference
  on Medical Image Computing and Computer-Assisted Intervention (MICCAI)},
  2015, pp. 234--241.

\bibitem{chen2018learning}
C.~Chen, Q.~Chen, J.~Xu, and V.~Koltun, ``Learning to see in the dark,'' in
  \emph{Proc. IEEE Conference on Computer Vision and Pattern Recognition
  (CVPR)}, 2018, pp. 3291--3300.

\bibitem{jaroensri2019generating}
R.~Jaroensri, C.~Biscarrat, M.~Aittala, and F.~Durand, ``Generating training
  data for denoising real rgb images via camera pipeline simulation,''
  \emph{arXiv}, 2019.

\bibitem{mao2016image}
X.~Mao, C.~Shen, and Y.-B. Yang, ``Image restoration using very deep
  convolutional encoder-decoder networks with symmetric skip connections,'' in
  \emph{Advances in Neural Information Processing Systems (NIPS)}, 2016, pp.
  2802--2810.

\bibitem{chen2016trainable}
Y.~Chen and T.~Pock, ``Trainable nonlinear reaction diffusion: A flexible
  framework for fast and effective image restoration,'' \emph{IEEE Transactions
  on Pattern Analysis and Machine Intelligence}, vol.~39, no.~6, pp.
  1256--1272, 2016.

\bibitem{he2015delving}
K.~He, X.~Zhang, S.~Ren, and J.~Sun, ``Delving deep into rectifiers: Surpassing
  human-level performance on imagenet classification,'' in \emph{Proc. IEEE
  International Conference on Computer Vision (ICCV)}, 2015, pp. 1026--1034.

\bibitem{chen2017deeplab}
L.-C. Chen, G.~Papandreou, I.~Kokkinos, K.~Murphy, and A.~L. Yuille, ``Deeplab:
  Semantic image segmentation with deep convolutional nets, atrous convolution,
  and fully connected crfs,'' \emph{IEEE Transactions on Pattern Analysis and
  Machine Intelligence}, vol.~40, no.~4, pp. 834--848, 2017.

\bibitem{xiao2018unified}
T.~Xiao, Y.~Liu, B.~Zhou, Y.~Jiang, and J.~Sun, ``Unified perceptual parsing
  for scene understanding,'' in \emph{Proc. European Conference on Computer
  Vision (ECCV)}, 2018, pp. 418--434.

\bibitem{zhou2017scene}
B.~Zhou, H.~Zhao, X.~Puig, S.~Fidler, A.~Barriuso, and A.~Torralba, ``Scene
  parsing through ade20k dataset,'' in \emph{Proc. IEEE Conference on Computer
  Vision and Pattern Recognition (CVPR)}, 2017.

\bibitem{zhao2017pyramid}
H.~Zhao, J.~Shi, X.~Qi, X.~Wang, and J.~Jia, ``Pyramid scene parsing network,''
  in \emph{Proc. IEEE Conference on Computer Vision and Pattern Recognition
  (CVPR)}, 2017, pp. 2881--2890.

\bibitem{lin2017feature}
T.-Y. Lin, P.~Doll{\'a}r, R.~Girshick, K.~He, B.~Hariharan, and S.~Belongie,
  ``Feature pyramid networks for object detection,'' in \emph{Proc. IEEE
  Conference on Computer Vision and Pattern Recognition (CVPR)}, 2017, pp.
  2117--2125.

\bibitem{liu2015learning}
F.~Liu, C.~Shen, G.~Lin, and I.~Reid, ``Learning depth from single monocular
  images using deep convolutional neural fields,'' \emph{IEEE Transactions on
  Pattern Analysis and Machine Intelligence}, vol.~38, no.~10, pp. 2024--2039,
  2015.

\bibitem{xian2018monocular}
K.~Xian, C.~Shen, Z.~Cao, H.~Lu, Y.~Xiao, R.~Li, and Z.~Luo, ``Monocular
  relative depth perception with web stereo data supervision,'' in \emph{Proc.
  IEEE Conference on Computer Vision and Pattern Recognition (CVPR)}, 2018, pp.
  311--320.

\bibitem{wofk2019fastdepth}
D.~Wofk, F.~Ma, T.-J. Yang, S.~Karaman, and V.~Sze, ``Fastdepth: Fast monocular
  depth estimation on embedded systems,'' in \emph{Proc. IEEE International
  Conference on Robotics and Automation (ICRA)}, 2019.

\bibitem{hu2019revisiting}
J.~Hu, M.~Ozay, Y.~Zhang, and T.~Okatani, ``Revisiting single image depth
  estimation: toward higher resolution maps with accurate object boundaries,''
  in \emph{Proc. IEEE Winter Conference on Applications of Computer Vision
  (WACV)}, 2019, pp. 1043--1051.

\end{thebibliography}
\end{document}